\documentclass[acmtog,screen]{acmart}
\acmSubmissionID{1234}

\usepackage{booktabs} %

\usepackage[ruled]{algorithm2e} %

\SetAlFnt{\small}
\SetAlCapFnt{\small}
\SetAlCapNameFnt{\small}
\SetAlCapHSkip{0pt}

\acmJournal{TOG}

\usepackage{colortbl}
\usepackage{graphicx}
\usepackage{amsmath}
\usepackage{booktabs}
\usepackage{multirow}
\usepackage{cuted}
\usepackage{bm}
\usepackage{xspace}
\usepackage{caption}
\usepackage{balance}
\usepackage{pifont}
\usepackage[normalem]{ulem}
\usepackage{arydshln}
\usepackage{lipsum}
\usepackage{acronym} %
\usepackage{setspace}
\usepackage{attrib}
\usepackage{epigraph}

\usepackage{setspace}
\usepackage{enumitem}
\usepackage{dsfont}
\usepackage[capitalize]{cleveref}

\newcommand{\model}{DELTA\xspace}

\newcommand{\vect}[1]{\mathbf{#1}}
\newcommand{\norm}[1]{\left\lVert#1\right\rVert}

\newcommand{\shapecoeff}{\bm{\beta}}
\newcommand{\shapedim}{{\left| \shapecoeff \right|}}
\newcommand{\shapespace}{\mathcal{S}}
\newcommand{\shapespaceexpl}{\mathbb{R}^{\shapedim}}

\newcommand{\numjoints}{n_k}
\newcommand{\joints}{{J}}

\newcommand{\posecoeff}{\bm{\theta}}
\newcommand{\posedim}{{3\numjoints+3}}
\newcommand{\posespace}{\mathcal{P}}
\newcommand{\posespaceexpl}{\mathbb{R}^{\posedim}}

\newcommand{\expcoeff}{\bm{\psi}}
\newcommand{\expdim}{{\left| \expcoeff \right|}}
\newcommand{\expspace}{\mathcal{E}}
\newcommand{\expspaceexpl}{\mathbb{R}^{\expdim}}

\newcommand{\numverts}{n_v}
\newcommand{\numfaces}{n_t}
\newcommand{\template}{\bar{\bm{T}}}

\newcommand{\lbs}{\text{LBS}}

\newcommand{\vsmplx}{M}

\newcommand{\offsets}{{O}}
\newcommand{\offset}{\bm{o}}

\newcommand{\offsetmodel}{{F_{d}}}
\newcommand{\texmodel}{{F_{t}}}

\newcommand{\defmodel}{{F_{e}}}

\newcommand{\image}{I}
\newcommand{\imagemask}{S} %
\newcommand{\numframes}{n_f}
\newcommand{\framenum}{f}
\newcommand{\cam}{\vect{p}}
\newcommand{\meshrender}{\mathcal{R}_m}
\newcommand{\lossweight}[1]{\lambda_{#1}}

\newcommand{\volrender}{\mathcal{R}_v}
\newcommand{\col}{\bm{c}}

\newcommand{\density}{\sigma}

\newcommand{\smplx}{\mbox{SMPL-X}\xspace}

\newcommand{\hairmodel}{{F_{h}}}

\definecolor{Gray}{gray}{0.92}

\newcolumntype{a}{>{\columncolor{Gray}}c}

\newlength\savewidth\newcommand\shline{\noalign{\global\savewidth\arrayrulewidth
  \global\arrayrulewidth 1pt}\hline\noalign{\global\arrayrulewidth\savewidth}}

\newcommand{\ie}{\emph{i.e.}}
\newcommand{\eg}{\emph{e.g.}}

\begin{document}
\title{Learning Disentangled Avatars with Hybrid 3D Representations}

\author{Yao Feng}
\affiliation{%
  \institution{Max Planck Institute for Intelligent Systems, Germany \& ETH Zürich, Switzerland}
}
\author{Weiyang Liu}
\affiliation{%
  \institution{Max Planck Institute for Intelligent Systems, Germany \& University of Cambridge, United Kingdom}
}

\author{Timo Bolkart}
\affiliation{%
  \institution{Max Planck Institute for Intelligent Systems, Germany}
}
\author{Jinlong Yang}
\affiliation{%
  \institution{Max Planck Institute for Intelligent Systems, Germany}
}
\author{Marc Pollefeys}
\affiliation{%
  \institution{ETH Zürich, Switzerland}
}
\author{Michael J. Black}
\affiliation{%
  \institution{Max Planck Institute for Intelligent Systems, Germany}
}

\begin{abstract}
\textbf{Abstract:}~Tremendous efforts have been made to learn animatable and photorealistic human avatars. Towards this end, both explicit and implicit 3D representations are heavily studied for a holistic modeling and capture of the whole human (\eg, body, clothing, face and hair), but neither representation is an optimal choice in terms of representation efficacy since different parts of the human avatar have different modeling desiderata. For example, meshes are generally not suitable for modeling clothing and hair. Motivated by this, we present Disentangled Avatars~(DELTA), which models humans with hybrid explicit-implicit 3D representations. DELTA takes a monocular RGB video as input, and produces a human avatar with separate body and clothing/hair layers.
Specifically, we demonstrate two important applications for DELTA. For the first one, we consider the disentanglement of the human body and clothing and in the second, we disentangle the face and hair. To do so, DELTA represents the body or face with an explicit mesh-based parametric 3D model and the clothing or hair with an implicit neural radiance field. To make this possible, we design an end-to-end differentiable renderer that integrates meshes into volumetric rendering, enabling DELTA to learn directly from monocular videos without any 3D supervision.
Finally, we show that how these two applications can be easily combined to model full-body avatars, such that the hair, face, body and clothing can be fully disentangled yet jointly rendered. Such a disentanglement enables hair and clothing transfer to arbitrary body shapes.
We empirically validate the effectiveness of DELTA's disentanglement by demonstrating its promising performance on disentangled reconstruction, virtual clothing try-on and hairstyle transfer. To facilitate future research, we also release an open-sourced pipeline for the study of hybrid human avatar modeling. 
\end{abstract}

\begin{teaserfigure}
\vbox{%
         \vskip 0.05in
         \hsize\textwidth
         \linewidth\hsize
         \centering
         \normalsize
         {Project Page:}~\tt\href{https://yfeng95.github.io/delta}{yfeng95.github.io/delta}
         \vskip 0.05in
}
\end{teaserfigure}

\begin{teaserfigure}
  \includegraphics[width=1.00\linewidth]{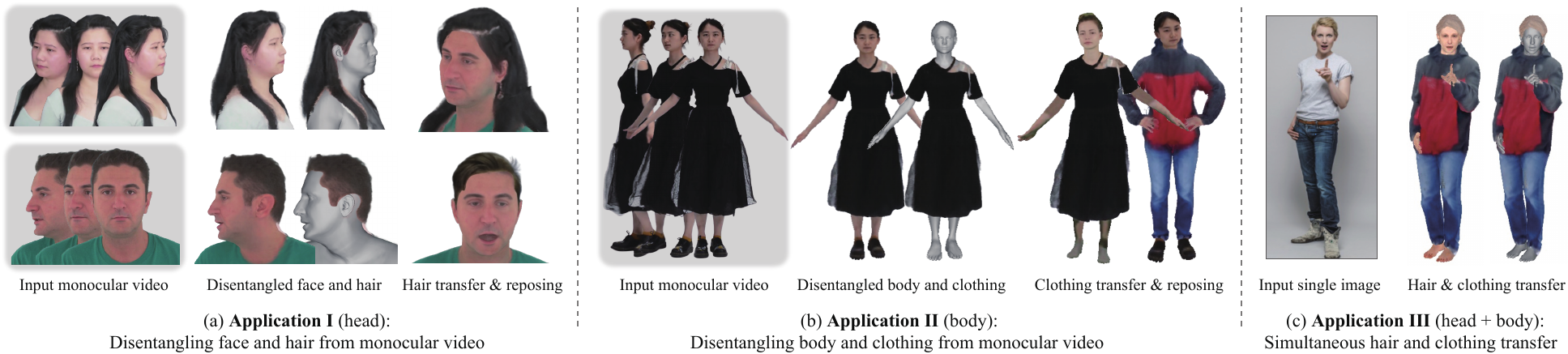}
  \vspace{-5mm}
  \caption{(a) Disentangled human head: \model outputs disentangled mesh-based face and NeRF-based hair given a monocular video input. (b) Disentangled human body: \model outputs disentangled mesh-based body and NeRF-based clothing given a monocular video input. (c) With the disentangled clothing and hair learned by \model, we can easily transfer any hair and clothing to a human body estimated from a single image. }
    \label{fig:teaser}
    \vspace{6mm}
\end{teaserfigure}

\maketitle

\section{Introduction}
\label{sec:intro}
Recent years have witnessed an unparalleled surge in the utilization of 3D human reconstruction and reenactment in numerous applications such as virtual and augmented reality, telepresence, games, and movies. It is of broad interest to create personal avatars from readily available setups (\eg, monocular videos). It is desirable in practice for the avatars to be photorealistic, 3D-consistent, animatable, easily editable and generalizable to novel poses. These characteristics call for a faithful disentanglement and modeling of different semantic components of the avatar (\eg, face and hair for head, body and clothing for whole body). Therefore, how to disentangle human avatars while yielding accurate reconstructions is of great significance and remains an open challenge.

Existing methods for learning 3D human avatars can be roughly categorized into \emph{explicit} ones and \emph{implicit} ones. Explicit methods (\eg, \cite{sanyal2019learning,feng2021learning,grassal2022neural,khakhulin2022realistic} for head, \cite{Choutas2020ExPose,feng2021collaborative,kanazawa2018end,kolotouros2019learning,Pavlakos2019_smplifyx,zanfir2021neural} for body) typically use triangular meshes as representation, and the reconstruction heavily relies on statistical shape priors, such as 3D morphable models for head~\cite{blanz1999morphable,li2017learning,egger20203d} and 3D parametric models for body~\cite{anguelov2005scape,SMPL:2015,Pavlakos2019_smplifyx,xu2020ghum,joo2018total,osman2020star}. Implicit methods usually encode the 3D geometry either with implicit surfaces (\eg, signed distance fields~(SDF))~\cite{zheng2022avatar,saito2019pifu,jiang2022selfrecon} or with volumetric representation~\cite{gafni2021dynamic,Gao2022nerfblendshape,peng2021neural}. Both explicit and implicit methods use a single 3D representation to model different parts of the avatar, which ignores the representation efficacy and therefore can be sub-optimal. For example, triangular meshes are an efficient representation for faces and minimally clothed body, for which statistical template priors are available, but meshes are generally a poor representation for hair or clothing since they can be inefficient to capture the underlying geometry. On the other hand, implicit representation renders high-fidelity 2D views but it is nontrivial to animate and usually can not generalize to unseen poses and expressions. Since no single 3D representation is perfect, \emph{why not use different one for different part of the avatar?} Motivated by this, we propose \textbf{D}is\textbf{E}ntang\textbf{L}ed ava\textbf{TA}r~(\model), which models face and body with explicit triangular meshes, and models hair and clothing with an implicit neural radiance field~(NeRF)~\cite{mildenhall2020nerf}. The intuition behind such a design is in two folds. First, both faces and bodies have regular topological structures and live in a low-dimensional subspace~\cite{basri2003lambertian,li2009expression}. It is therefore a well-motivated choice to represent the face or body geometry with mesh templates. Second, hair consists of countless freely deformed thin strands, which hinders triangular meshes to be a suitable representation. Clothing (\eg, dresses) also consists of complex topological structures and has a diverse set of styles. Due to the complex nature of hair and clothing, it is highly difficult to accurately model their surface geometry, which renders NeRF an arguably better choice of representation.

The effectiveness of hybrid 3D representation has already found its traces in human-scene reconstruction~\cite{pavlakos2022one}, clothed body modeling~\cite{feng2022scarf}, and human eye modeling~\cite{li2022eyenerf}. For example, \cite{pavlakos2022one} reconstructs the static scene with a NeRF which excels at representing fine-grained scene details, and the people inside with a SMPL~\cite{SMPL:2015} representation which is good at body pose recovery. Despite modeling different subjects under different context, the essence of hybrid representation is the adoption of heterogeneous 3D representations such that each representation can be made the best use of. Extending our prior work~\cite{feng2022scarf}, \model is the \emph{first} method to demonstrate the power of hybrid representation for learning human avatars (including face, body, hair and clothing). Specifically, we instantiate the idea of \model in two capture settings. First, we consider the disentangled reconstruction of human head where the head (and upper shoulder) is represented by a parametric mesh model (\ie, FLAME~\cite{li2017learning} and SMPL-X~\cite{Pavlakos2019_smplifyx}) and the hair is represented by a NeRF. Unlike existing works~\cite{gafni2021dynamic,grassal2022neural,zheng2022avatar}, DELTA additionally reconstruct the upper body (\eg, shoulder), such that people with long hair can be better captured. Second, we consider the disentangled reconstruction of human body where the body is represented by a parametric mesh model (\ie, SMPL-X) and the clothing is represented by a NeRF. Combining the disentangled capture of both human head and body, we demonstrate that both hair and clothing can be simultaneously transferred to arbitrary reconstructed human body. See Figure~\ref{fig:teaser} for an illustration.

Distinct from existing work~\cite{pavlakos2022one,li2022eyenerf}, at the very heart of \model is our novel mesh-integrated volumetric renderer, which not only drives the disentanglement of different parts of the avatar (\ie, face, hair, body, clothing), but also enables the end-to-end differentiable learning directly from monocular videos without any 3D supervision. We expect the idea of hybrid 3D representation to be quite general, and \model aims to demonstrate the power of hybrid 3D representation by bringing together meshes and NeRFs in modeling human avatars.

\emph{Why is disentanglement so important for learning avatars?} 
We answer this question by listing some key desiderata for photorealistic avatar creation. First, the pose-dependent factors should be disentangled from the appearance such that the captured avatar can be easily reusable in new environments. 
Second, disentangling the human body, hair, and clothing is crucial to accurately model their respective dynamics, since the motion dynamics of the human body, hair, and clothing are completely distinct from each other. Moreover, modeling the interaction between body and hair/clothing also requires an accurate disentanglement. Such a disentanglement becomes even more important when performing physical simulation on the reconstructed avatar. 
Third, human body, hair and clothing have totally different material and physical properties, which results in different lighting phenomena. In order to construct realistic and generalizable avatars, human body and hair/clothing have to be disentangled and modeled separately. Towards the goal of learning disentangled avatars, our contributions are listed below: 

\vspace{0.7mm}
\begin{itemize}[leftmargin=*,nosep]
\setlength\itemsep{0.3em}
    \item By substantially extending our previous work~\cite{feng2022scarf}, we propose the disentangled avatar that models face/body and hair/clothing with a hybrid 3D representation. Such an hybrid representation marries the statistical prior from mesh surfaces and the representation flexibility from implicit functions. \model is one of the first methods that uses a hybrid explicit-implicit representation to reconstruct high-fidelity disentangled avatars.
    \item We design a novel differentiable volumetric rendering method that incorporates meshes into volumetric rendering.
    \item The framework of \model is fully differentiable and end-to-end trainable. It is trained on a monocular video (\eg, from web cameras) without requiring any 3D supervision.
    \item For the face and body, \model delivers high-fidelity details while being able to effortlessly reposed. For the hair and clothing region, \model yields realistic hair and clothing reconstruction owing to the powerful implicit NeRF representation.
    \item We emphasize that the major contribution of \model is to serve as a demonstration to showcase the potentials of hybrid 3D representation in modeling human avatars.  
\end{itemize}

\section{Related Work}
\label{sec:relatedwork}

\subsection{Head Avatar Creation}
\textbf{Explicit head avatars}. 
Explicit head avatars are typically based on explicit 3D representations (\eg, triangular meshes). 3D morphable models (3DMM)~\cite{blanz1999morphable}, which are obtained from a population of 3D head scans~\cite{egger20203d}, are widely used as a stronger statistical prior to represent the geometry of faces. Built upon 3DMM, many improved variants have been proposed, including multi-linear models for shape and expression~\cite{cao2013facewarehouse,vlasic2006face}, full-head models~\cite{dai2020statistical,li2017learning,ploumpis2020towards}, and deep nonlinear models~\cite{ranjan2018generating,tran2018nonlinear}. Besides, morphable models also provide a linear model for textures~\cite{aldrian2010linear,blanz1999morphable,blanz2003face,paysan20093d}. 3DMM and its variants can be used to reconstruct faces through an optimization procedure~\cite{gecer2019ganfit,romdhani2005estimating,schonborn2017markov,thies2016face2face} or learning-based estimation~\cite{deng2019accurate,dib2021towards,feng2021learning,lattas2020avatarme,khakhulin2022realistic,li2018feature,sanyal2019learning,shang2020self,wen2021self,tewari2019fml,tewari2018self,tewari2017mofa}. Besides 3DMM template priors, other priors (\eg, symmetry~\cite{wu2020unsupervised,liu2022structural}, causality~\cite{liu2022structural,wen2021self}, identity~\cite{cole2017synthesizing,feng2021learning}) are also considered in 3D face reconstruction. 
Despite producing good coarse facial geometry, these methods are usually unable to reconstruct fine-grained facial details and the entire head (\eg, hair). 
Some methods~\cite{alldieck2018detailed,cao2015real,feng2021learning} use mesh displacements to reconstruct fine details such as wrinkles, producing fine-grained geometry. Following a similar spirit, \citet{grassal2022neural} use a geometry refinement network that learns a pose-dependent offset function for geometry corrections, and produces photorealistic outputs under novel views. PointAvatar~\cite{Zheng2023pointavatar} uses a deformable point-based representation to reconstruct human heads from videos.  
Unlike previous work, \model captures the head avatar with disentangled face and hair components. \model adopts the explicit mesh-based representation to model the face region, making it easily animatable. For the hair, we utilize an implicit NeRF-based representation, capable of accommodating various hair types. With this approach, we can utilize models tailored for faces and hair, and it also unlocks potential applications like hairstyle transfer.

\noindent\textbf{Implicit head avatars}. 
Implicit models normally encode the 3D head avatar with NeRF-based  representation~\cite{mildenhall2020nerf,mueller2022instant} or implicit surface functions~\cite{chen2019learning,kellnhofer2021neural,mescheder2019occupancy,park2019deepsdf,yariv2020multiview}. 
NeRF-based methods have been explored for 3D face modeling from images or videos~\cite{chan2021pi,gafni2021dynamic,wang2021learning,park2021nerfies}. 
\citet{gafni2021dynamic} reconstruct an animatable NeRF from a single monocular video, which is conditioned on the expression code from a 3DMM. \citet{Gao2022nerfblendshape} propose a NeRF-based linear blending representation where expression is encoded by multi-level voxel fields. AvatarMAV~\cite{xu2023avatarmav} uses neural voxel fields to represent motion and appearance to achieve fast head reconstruction. LatentAvatar~\cite{xu2023latentavatar} reconstructs a NeRF-based head avatar that is driven by latent expression codes, and these expression codes are learned in an end-to-end and self-supervised manner without the tracking of templates.
However, NeRF-based head representations generally suffer from poor 3D geometry and struggles to generalize to unseen poses/expressions. 
Approaches utilizing implicit surface functions generally provide better geometry for faces. 
\citet{yenamandra2021i3dmm} proposes an implicit morphable face model that disentangles texture and geometry. \citet{zheng2022avatar} parameterize the head with implicit surface functions in the canonical space, and represents the expression- and pose-dependent deformations via learned blendshapes and skinning fields. \citet{ramon2021h3d} use an optimization-based approach to estimate the signed distance function (SDF) of a full head from a few images, and this optimization is constrained by a pre-trained 3D head SDF model. 
In contrast to both explicit and implicit head avatars that use a holistic 3D representation, \model is the first method that adopts a hybrid explicit-implicit 3D representation to separately model face and hair. \model marries the strong controllability of the mesh-based face and the high-fidelity rendering of the NeRF-based hair. 

\subsection{Full Body Avatar Creation}
\noindent\textbf{Explicit Body Avatars}. The 3D surface of a human body is typically represented by a learned statistical 3D model using an explicit mesh representation~\cite{anguelov2005scape, joo2018total, SMPL:2015, osman2020star, Pavlakos2019_smplifyx}. The parametric models~\cite{Pavlakos2019_smplifyx, SMPL:2015} can produce a minimal clothed body when the shape parameters are provided. 
Numerous optimization and regression methods have been proposed to compute 3D shape and pose parameters from images, videos, and scans.
See \cite{tian2022hmrsurvey, Liu2021hmrsurvey} for recent surveys.
We focus on methods that capture full-body pose and shape, including the hands and facial expressions \cite{Pavlakos2019_smplifyx, Choutas2020ExPose, feng2021collaborative, Xiang2019, Rong2021Frankmocap, Zhou2021, xu2020ghum}. Such methods, however, do not capture hair, clothing, or anything that deviates the body. Also, they rarely recover texture information, due to the large geometric discrepancy between the clothed human in the image and captured minimal clothed body mesh. 
Some methods choose to model body along with clothing. However, clothing is more complex than the body in terms of geometry, non-rigid deformation, and appearance, making the capture of clothing from images challenging.  
Explicit ways to capture clothing often use additional vertex offsets relative to the body mesh \cite{alldieck2018video, alldieck2018detailed, lazova3dv2019, alldieck2019learning, alldieck2019tex2shape,  ma2020learning, jin2020pixel, xiu2023econ}. 
While such an approach generally works well for tight clothing, it still struggles to capture loose clothing like skirts and dresses.  

\vspace{0.75mm}
\noindent\textbf{Implicit Body Avatars}. Recently, implicit representations have gained traction in modeling the human body~\cite{xu2020ghum, Alldieck2021_imGHUM}. Correspondingly, methods have been developed to estimate implicit body shape from images~\cite{xu2020ghum}. However, similar to explicit body model~\cite{Pavlakos2019_smplifyx}, they only model minimal clothed body. 
When it comes to clothed avatars, recent methods are leveraging implicit representations to handle more complex variations in clothing styles, aiding in the recovery of clothing structures. 
For instance, \cite{huang2020arch, he2021arch++, saito2019pifu, saito2020pifuhd, xiu2022icon, zheng2021pamir} extract pixel-aligned spatial features from images and map them to an implicit shape representation.
To animate the captured non-parametric clothed humans, 
\citet{yang2021s3} predict skeleton and skinning weights from images to drive the representation.
\citet{corona2021smplicit} represent clothing layers with deep unsigned distance functions~\cite{chibane2020ndf}, and learn the clothing style and clothing cut space with an auto-decoder. 
Once trained, the clothing latent code can be optimized to match image observations, but it produces over-smooth results without detailed wrinkles. PoseVocab~\cite{li2023posevocab} models NeRF-based human avatars by learning pose encoding.
Although such implicit models can capture various clothing styles much better than explicit mesh-based approaches, faces and hands are usually poorly recovered due to the lack of a strong prior on the human body. In addition, such approaches typically require a large set of manually cleaned 3D scans as training data. 
Recently, various methods recover 3D clothed humans directly from multi-view or monocular RGB videos \cite{su2021nerf, weng2022humannerf, liu2021neural, peng2021neural, chen2021animatable, peng2021animatable, jiang2022selfrecon, peng2022animatable, qiu2023rec}. They optimize avatars from image information using implicit shape rendering \cite{liu2020dist, yariv2020multiview, yariv2021volume, niemeyer2020differentiable} or volume rendering~\cite{mildenhall2020nerf}, no 3D scans are needed. 
Although these approaches demonstrate impressive performance, hand gestures and facial expressions are difficult to capture and animate due to the lack of model expressiveness and controllability. AvatarReX~\cite{zheng2023avatarrex} learns a NeRF-based full-body avatar with disentangled modeling of face, body and hands, but the clothing is still entangled with body.

Unlike prior methods, we view clothing as a separate layer above the body and combine explicit body models and implicit clothing to leverage the advantages of both. 
The mesh-based body model allows us to create human shapes with detailed components (\eg, hands) and to control the body (\eg, expressions and hand articulations). With implicit representation, we can capture a variety of clothing using images, without the need for 3D scans.
Moreover, the disentangled modeling of explicit body and implicit clothing facilitates seamless clothing transfer, enabling applications like virtual try-ons.

\subsection{Other Related Work}

\textbf{Hybrid 3D representation}. The potentials of hybrid 3D representation have also been demonstrated in other 3D reconstruction tasks. \citet{pavlakos2022one} represent the background static scene as a NeRF and the people inside as SMPL models. \citet{li2022eyenerf} model the eye-ball surface with an explicit parametric surface model and represents the periocular region and the interior of the eye with deformable volumetric representations. Hybrid explicit-implicit representation has also been explored in transparent object reconstruction~\cite{xu2022hybrid} and haptic rendering~\cite{kim2004haptic}.

\vspace{0.5mm}
\noindent\textbf{Hair modeling}. How to represent hair is a long-standing problem in human modeling~\cite{ward2007survey}. Strand-based modeling is widely adopted to model human hair~\cite{beeler2012coupled,luo2012multi,luo2013structure,hu2014robust,herrera2012lighting,nam2019strand,sun2021human,chai2013dynamic,chai2012single,zhang2017data,yang2019dynamic,zhang2019hair,zhou2018hairnet,rosu2022neural}. \citet{zheng2023hairstep} recover the strand-based 3D hair from an intermediate representation that consists of a strand map and a depth map. Neural Haircut~\cite{sklyarova2023neural_haircut} uses a two-stage coarse-to-fine optimization to reconstruct the strand-level hair.  More recently, volumetric representation is also applied to perform hair modeling~\cite{saito20183d,wang2022hvh}. 
Their primary focus is on hair reconstruction, and they typically utilize head-tracked meshes from multi-view images \cite{wang2022hvh, wang2021learning, rosu2022neural} or reconstruct faces from videos with stationary heads \cite{sklyarova2023neural_haircut}. None of these methods, however, are designed to learn faces from monocular videos with dynamic facial expressions.
In contrast, our approach distinguishes itself by learning both facial features and hair from monocular videos, even when the head is moving. Since the primary objective of \model is to disentangle the representation of faces and hair rather than accurately capturing hair geometry, we employ a NeRF representation for hair modeling.
The disentangled capture of face, upper body and hair is a necessary step before one can perform high-fidelity hair modeling, so \model also serves as a stepping stone for future work that combines better hair modeling in creating disentangled head avatars.

\noindent\textbf{Garment reconstruction}. 
The task of reconstructing 3D garments from images or videos has proven to be a complex challenge \cite{zhu2020deep, hong2021garment4d, zhao2021learning, danvevrek2017deepgarment, qiu2023rec, su2022deepcloth, li2021deep}. This complexity arises from the wide diversity in clothing topologies.
To tackle this, existing methods often rely on either clothing template meshes or implicit surface functions. Typically, these approaches demand access to 3D data.   
Many approaches employ training data produced by physics-based simulations~\cite{bertiche2020cloth3d, santesteban2019learning, vidaurre2020fully, patel2020tailornet} or require template meshes fit to  3D scans \cite{pons2017clothcap, xiang2021modeling, tiwari2020sizer,chen2021tightcap,halimi2022garment}. 
\citet{jiang2020bcnet} train a mesh-based multi-clothing model on 3D datasets with various clothing styles. \citet{zhu2020deep} introduce a adaptable template that allows for encoding clothing with diverse topologies within a single mesh template. 
Then during inference, a trained network produces the 3D clothing as a separate mesh-based layer by recognizing and predicting the clothing style from an image. 
\citet{zhu2022registering} fit template meshes to non-parametric 3D reconstructions. 
While these methods recover garments from images, they are limited in visual fidelity, as they do not capture clothing appearance.
Additionally, methods with such predefined clothing style templates can not easily handle the real clothing variations, limiting their applications. 
In contrast, \citet{corona2021smplicit} represent clothing layers with deep unsigned distance functions \cite{chibane2020ndf}, and learn the clothing style and clothing cut space with an auto-decoder. 
Once trained, the clothing latent code can be optimized to match image observations, but it produces over-smooth results without detailed wrinkles. 
Instead, \model models the clothing layer with a neural radiance field, and optimizes the body and clothing layer from scratch instead of the latent space of a learned clothing model. 
Therefore, \model produces avatars with higher visual fidelity (see Section~\ref{sec:experiments}).

\section{\textbf{DELTA}: Learning Disentangled Avatars}
\begin{figure*}[t]
	\centering
	\includegraphics[width=\textwidth, trim={0cm, 0cm, 0cm, 0cm}]{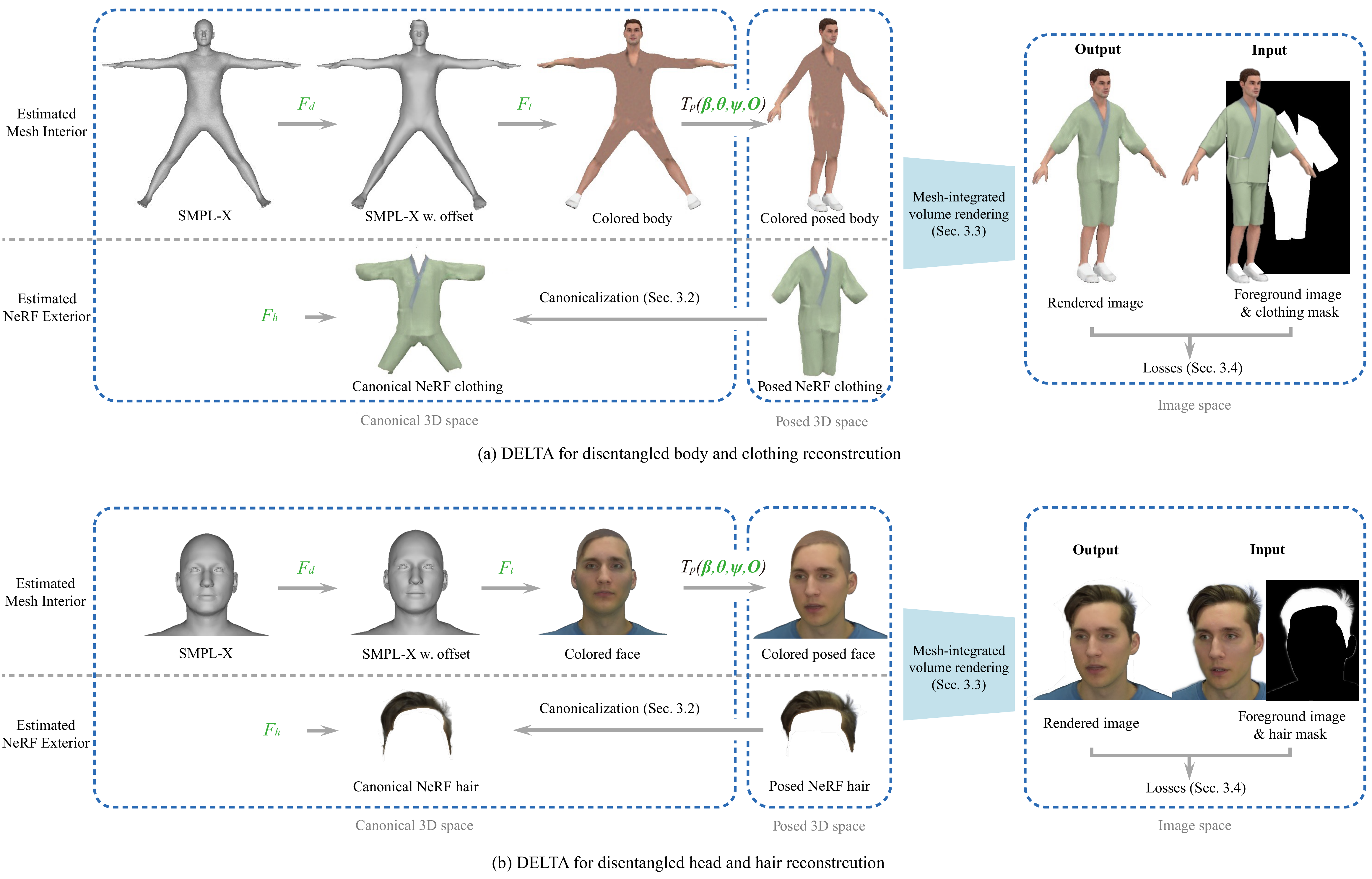}  
  \vspace{-5.5mm}
	\caption{\model takes a monocular RGB video and clothing/hair segmentation masks as input, and outputs a human avatar with separate body and clothing/hair layers. Green letters indicate optimizable modules or parameters.}
	\label{fig:pipeline}
 \vspace{1.5mm}
\end{figure*}

Given a monocular video, \model reconstructs a head (or body) avatar where head/body and hair/clothing are fully disentangled. Once the avatar is built, we can animate it with novel poses and change the hairstyle and clothing effortlessly. 
Because the way that \model reconstructs head and body shares many similarities, 
we simplify the description by  referring the face or body as \emph{avatar interior} and the hair or clothing as \emph{avatar exterior}.

\subsection{Hybrid Explicit-Implicit 3D Representations}
Previous work on face and body modeling~\cite{lombardi2018deep, bi2021deep, grassal2022neural,SMPL:2015,li2017learning,Pavlakos2019_smplifyx} has demonstrated that both human faces and bodies can be accurately modeled by mesh-based representations. In the light of these encouraging results, we choose mesh as the representation for the face and body. Specifically, we use \smplx~\cite{Pavlakos2019_smplifyx} to make full use of the human geometry priors. When it comes to representing hair and clothing, it remains an open problem which representation works the best. Because of the complex geometry of hair and clothing, we propose to model both hair and clothing with NeRF~\cite{mildenhall2020nerf} -- a more flexible and expressive implicit representation. 
Distinct from meshes, NeRF is agnostic to the style, geometry and topology of hair and clothing. 

\vspace{0.75mm}

\noindent\textbf{Explicit avatar interior by \smplx}.
\smplx is an expressive body model with detailed face shape and expressions. A subject's face and body with neutral expression in the rest pose is defined as
\begin{equation}
    T_P(\shapecoeff,\posecoeff, \expcoeff) = \template + B_{S}(\shapecoeff;\shapespace) +  B_{P}(\posecoeff;\posespace) + B_{E}(\expcoeff;\expspace),
\end{equation}
where $\template \in \mathbb{R}^{\numverts \times 3}$ is a template of body shape in the rest pose, $\shapecoeff \in \shapespaceexpl$ is the body identity parameters, and $B_{S}(\shapecoeff;\shapespace): \shapespaceexpl \rightarrow \mathbb{R}^{\numverts \times 3}$ are the identity blend shapes. More specifically, $B_S(\shapecoeff;\shapespace)=\sum_{i=1}^{|\shapecoeff|}\shapecoeff_i\shapespace_i$ where $\shapecoeff_i$ is the $i$-th linear coefficient and $\shapespace_i$ is the $i$-th orthonormal principle component. $\posecoeff\in\posespaceexpl$ denotes the pose parameters, and $\expcoeff\in\expspaceexpl$ denotes the facial expression parameters. Similar to the shape space $\shapespace$, $B_{P}(\posecoeff;\posespace): \mathbb{R}^{|\posecoeff|} \rightarrow \mathbb{R}^{\numverts \times 3}$ denotes the pose blend shapes ($\posespace$ is the pose space), and $B_{E}(\expcoeff;\expspace): \expspaceexpl \rightarrow \mathbb{R}^{\numverts \times 3}$ denotes the expression blend shapes from the \smplx model ($\expspace$ is the expression space). To increase the flexibility of \smplx, we add additional vertex offsets $\bm{\offsets}:=\{F_d(\bm{t}_1), F_d(\bm{t}_2), \cdots, F_d(\bm{t}_{\numverts})\}^\top\in \mathbb{R}^{\numverts \times 3}$ in the canonical space. 
The offset is modeled by a vertex-wise implicit function $\thickmuskip=2mu \medmuskip=2mu \offsetmodel: \bm{t} \rightarrow \offset$, which predicts an offset $\offset\in\mathbb{R}^3$ for the vertex $\bm{t}\in\mathbb{R}^3$ in the rest template. 
Therefore, we augment the body shape with the following set of offsets:
\begin{equation}
\begin{aligned}
    \tilde{T}_P(\shapecoeff, \posecoeff, \expcoeff, \bm{O}) = T_P(\shapecoeff,\posecoeff, \expcoeff) + \bm{O}.
\end{aligned}
\end{equation}

The albedo is represented by an implicit function $\texmodel: \bm{t} \rightarrow \bm{c}^{\text{mesh}}$ which predicts the RGB color $\bm{c}^{\text{mesh}}$ of each given vertex $\bm{t}$ on the surface. Specifically, we sample vertex $\bm{t}$ from the template mesh $\template$ if the video is under uniform lighting. For more complex lighting conditions, in order to better model the texture, we sample $\bm{t}$ from the surface after the pose deformation. More details can be found in Section~\ref{sect:implementation}.
To capture more geometric details, we use an upsampled version of \smplx with $\numverts=38,703$ vertices and $\numfaces=77,336$ faces~\cite{feng2022scarf}. Similar to \cite{grassal2022neural}, we also add additional faces inside the mouth region for head avatar modeling.

\vspace{0.75mm}
\noindent\textbf{Implicit avatar exterior by NeRF}.
Based on NeRF~\cite{mildenhall2020nerf}, we define the avatar exterior (hair or clothing) in the canonical 3D space as an implicit function $\hairmodel: \bm{x}^{c} \rightarrow ({\bm{c}}^{\text{nerf}}, \density)$ which can be parameterized by a multi-layer perceptron~(MLP). $\bm{c}^{\text{nerf}}$ represents the RGB color. Given a query point  $\bm{x}^{c} \in \mathbb{R}^3$ in the canonical space, the implicit NeRF-based function $\hairmodel$ outputs an emitted RGB color $\bm{c}^{\text{nerf}}$ and a volume density $\density$.

\subsection{Pose-dependent Deformation} 
\textbf{Explicit avatar interior deformation}. Given the monocular video, we need to model the movement of this subject. 
Since our avatar interior model is based on \smplx, it provides a good way to capture the pose deformation and facial expressions. 
For each frame of given video, we estimate the parameters of shape  $\posecoeff \in \mathbb{R}^{|\posecoeff|}$ and expression $\expcoeff \in \expspaceexpl$. 
Then we can deform the head/body to the observation pose using the linear blend skinning function (\ie, $\lbs$).  
The deformation for the explicit SMPL-X mesh model is modeled by a differential function $\vsmplx(\shapecoeff, \posecoeff, \expcoeff, \bm{O})$ that outputs a 3D human body mesh $(\bm{V},\bm{F})$ where $\bm{V}\in\mathbb{R}^{\numverts\times 3}$ is a set of $\numverts$ vertices and $\bm{F}\in\mathbb{R}^{\numfaces\times 3}$ is a set of $\numfaces$ faces with a fixed topology:
\begin{equation}
        \vsmplx(\shapecoeff, \posecoeff, \expcoeff, \bm{O}) =  \lbs(\tilde{T}_P(\shapecoeff, \posecoeff, \expcoeff, \bm{O}), \joints(\shapecoeff), \posecoeff, \bm{W}),
\end{equation}
in which $\thickmuskip=2mu \medmuskip=2mu \bm{W} \in \mathbb{R}^{\numjoints \times \numverts}$ is the blend skinning weights used in the $\lbs$ function. $\thickmuskip=2mu \medmuskip=2mu\joints(\shapecoeff)\in\mathbb{R}^{n_k\times 3}$ is a function of body shape~\cite{Pavlakos2019_smplifyx}, representing the shape-dependent joints. Given a template vertex $\bm{t}_i$, the vertex $\bm{v}_i$ can be computed with simple linear transformation. Specifically, the forward vertex-wise deformation can be written as the following equation in the homogeneous coordinates:
\begin{equation}
\begin{aligned}
    \underbrace{\bm{v}_i}_{\textnormal{Posed vertex}} = \underbrace{\sum_{k=1}^{n_k}\bm{W}_{k,i}G_k(\bm{\theta},J(\bm{\beta}))\cdot
        \begin{bmatrix}
     \bm{I} &  \bm{o}_i+ \bm{b}_i \\
      \bm{0} &  1 
  \end{bmatrix}}_{\vsmplx_i(\bm{\beta},\bm{\theta},\bm{\psi},\bm{O})\textnormal{:~Deformation to the posed space}}\cdot\underbrace{\bm{t}_i}_{\textnormal{Template vertex}}\nonumber,
\end{aligned}
\end{equation}
where $\vsmplx_i(\bm{\beta},\bm{\theta},\bm{\psi},\bm{O}) \in \mathbb{R}^{4\times 4}$ is the deformation function of template vertex $\bm{t}_i$.   
$\bm{W}_{k,i}$ is the $(k,i)$-th element of the blend weight matrix $\bm{W}$, $G_k(\bm{\theta},J(\bm{\beta}))\in\mathbb{R}^{4\times 4}$ is the world transformation of the $k$-th joint and $\bm{b}_i$ is the $i$-th vertex of the sum of all blend shapes $\bm{B} := B_S(\bm{\beta})+B_P(\bm{\theta})+B_E(\bm{\psi})$. We denote $\bm{V}$ as the vertex set of the posed avatar ($\bm{v}_i\in\bm{V}$). Both $\bm{v}_i$ and $\bm{t}_i$ are the homogeneous coordinates when applying this deformation function. 

\vspace{0.5mm}
\noindent\textbf{Implicit avatar exterior deformation}. Aiming to learn the NeRF-based clothing/hair representation in the canonical space, we need to deform from the posed space to the canonical space. Therefore, we perform backward deformation on the top of the explicit body skinning. 
Given a query point $\bm{x}^p$ in the posed space (from the observed video frame), we first find the nearest $k$ points on the body surface $M$.  
Then we use the weighted backward skinning function to transform the posed point $\bm{x}^p$ to the canonical space (\ie, $x^c$). 
To model more accurate clothing/hair movement and deformation, we further learn a pose-dependent deformation function $\defmodel: (\bm{x}^c,\bm{v}^p_{n(\bm{x}^p)})\in\mathbb{R}^6 \rightarrow \Delta\bm{x}^c\in\mathbb{R}^3$, where $\bm{x}^p$ denotes a point in observation space and $n(\bm{x}^p)$ is the set of indices of the nearest points to $\bm{x}^p$ in $\bm{V}^p$ which denotes the posed body meshes in $M(\bm{0},\bm{\theta},\bm{0},\bm{0})$. $F_e$ aims to predict the detailed non-rigid deformation for the query point in the canonical space. Then the residual $\Delta\bm{x}^c$ is added back to $\bm{x}^c$, and the displaced point $\tilde{\bm{x}}^c=\bm{x}^c+\Delta\bm{x}^c$ is fed to the canonical NeRF model $F_h$ in order to compensate the exterior clothing/hair deformation in the observation space. Specifically, we have the inverse blend skinning mapping from the observation space to the posed space as the following transformation:
\begin{equation}
\begin{aligned}
    \underbrace{\bm{x}^c}_{\substack{\textnormal{Canonical}\\ \textnormal{ vertex}}}\!=\!\!\underbrace{\sum_{\bm{v}_i\in n(\bm{x}^p)} \!\!\!\!\alpha_i(\bm{x}^p)\!\cdot\!{M}_i(\bm{0},\bm{\theta},\bm{0},\bm{0})\!\cdot\!{M}^{-1}_i(\bm{\beta},\bm{\theta},\bm{\psi},\bm{O})}_{\textnormal{Transformation to the canonical space}}\cdot \!\underbrace{\bm{x}^p}_{\substack{\textnormal{Observed}\\ \textnormal{ vertex}}}\nonumber,
\end{aligned}
\end{equation}
where $\alpha_i$ is the parameter that balances the importance:
\begin{equation}
\begin{aligned}
\alpha_i(\bm{x}^p)= \frac{1}{Z}\exp\left( -\frac{1}{2\sigma^2}\cdot\|\bm{x}^p-\bm{v}_i\|\cdot\|\bm{w}_{nn(\bm{x}^p)}-\bm{w}_i\|\right) \nonumber.
\end{aligned}
\end{equation}
 Where $Z:=\sum_{\bm{v}_i\in n(\bm{x}^p)}\alpha_i(\bm{x}^p)$ is a normalizing coefficient, $\bm{w}_i\in\mathbb{R}^{n_k}$ is the blend weights of $\bm{v}_i$, $\sigma$ is a constant and $nn(\bm{x}^p)$ denotes the index of the nearest point of $\bm{x}^p$ in $\bm{V}^p$.
 
\subsection{Mesh-integrated Volume Rendering} 
\begin{figure}[t]
	\centering
	\includegraphics[width=\linewidth]{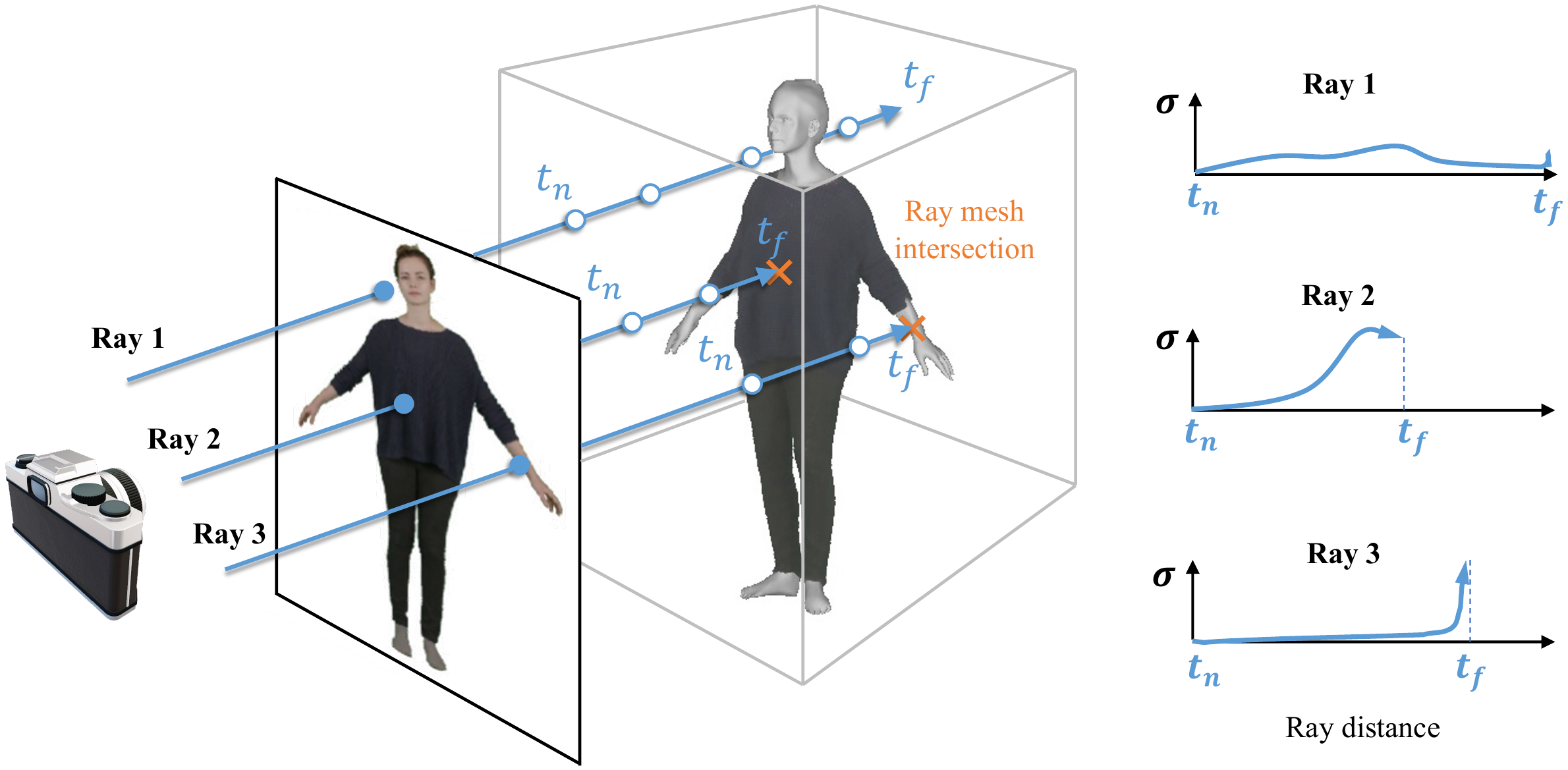}
	\caption{Illustration of mesh-integrated volume rendering.}
	\label{fig:rendering}
\end{figure}

\textbf{Camera model}. We simplify the problem by using a scaled orthographic camera model $\cam=\{s,\bm{t}^\top\}^\top$ where $s\in\mathbb{R}$ is the isotropic scale and $\bm{t}\in\mathbb{R}^2$ denotes the translation.

\vspace{0.75mm}

\noindent\textbf{Mesh rasterization}. With the geometry parameters ($\bm{\beta},\bm{\theta},\bm{\psi}$), the vertex offsets $\bm{O}$, the RGB color $\bm{c}^{\text{mesh}}$ of vertices in the upsampled SMPL-X template and the camera parameters $\cam$, we render the colored mesh into an image with $\mathcal{R}_m(\vsmplx(\shapecoeff, \posecoeff, \expcoeff, \offsetmodel),\bm{c}^{\text{mesh}},\cam)$ where $\mathcal{R}_m$ denotes the differentiable rasterizer function.

\vspace{0.75mm}

\noindent\textbf{Mesh-integrated volume rendering}. Finally we discuss how to take mesh into consideration while performing volumetric rendering.  
The basic idea is that the camera ray will stop when it intersects with the mesh in the 3D space. 
Given a camera ray $\bm{r}(t) = \bm{q} + t \bm{d}$ with center $\bm{q} \in \mathbb{R}^3$ and direction $\bm{d} \in \mathbb{R}^3$. The rendering interval is $t \in [t_n, t_f] \subset \mathbb{R}$ (near and far bounds).
Unlike previous work, we integrate the body model, 
$\vsmplx(\shapecoeff,\posecoeff,\expcoeff,\offsets)$, into the volumetric rendering. 
Specifically, if $\bm{r}(t)$ intersects $\vsmplx$, we set the $t_f$ such that $\bm{r}(t_f)$ is the intersection point with $\vsmplx$.
In this case, we use the mesh color instead of the NeRF color $\col^{\text{nerf}}(\bm{r}(t_f))$ (see Figure~\ref{fig:rendering}).
More formally, the expected color of the camera ray $r$ is defined as
\begin{equation}
\begin{aligned}
\bm{c}(\bm{r})=\int_{t_n}^{t_f}\bm{c}^{\text{nerf}}(\bm{r}(t))  \cdot T(t)\cdot\sigma(\bm{r}(t)) + \mathds{1}_{\text{s}}(\bm{r})\cdot\delta(t-t_f)\cdot\bm{c}^{\text{mesh}}dt\nonumber, 
\end{aligned}
\end{equation}
where $\mathds{1}_{\text{s}}(\bm{r})$ is the indicator function for whether the ray intersects the mesh surface ($1$ if true, $0$ otherwise), $\delta(\cdot)$ denotes the Dirac delta function and $T(t)=\exp(-\int_{t_n}^t\sigma(\bm{r}(s))ds)$. When $\mathds{1}_{\text{s}}(\bm{r})$ is true, we set the $t_f$ such that $\bm{t}(t_f)$ is the intersection point with the SMPL-X mesh $\vsmplx$.
$\bm{c}^{\text{mesh}}$ is the vertex color of the intersected mesh. We approximate the integral with evenly split $n_b$ bins in practice:
\begin{equation}
\begin{aligned}
    &\bm{c}(\bm{r})=\big(1-\sum_{k=1}^{n_b-1}T_k\big(1-\exp(-\sigma_k\Delta_k)\big)\big)\cdot\big((1-\mathds{1}_{\text{s}}(\bm{r})){\bm{c}}^{\text{nerf}}(\bm{r}^c_{n_b})\\[-1.5mm]
    &~~~~~~~~~~~~+\mathds{1}_{\text{s}}(\bm{r})\cdot\bm{c}^{\text{mesh}}(\bm{r}_{n_b})\big)+\sum_{j=1}^{n_b-1}T_j\big(1-\exp(-\sigma_j\Delta_j)\big){\bm{c}}^{\text{nerf}}(\bm{r}^c_{j})\nonumber,
\end{aligned}
\end{equation}
where we define $T_j=\exp(-\sum_{q=1}^{j-1}\sigma_j\Delta_j)$. $\bm{r}_j$ is sampled from the $j$-th bin along the camera ray $\bm{r}$. $\bm{r}^c_i$ is the corresponding canonical point for the observed point $\bm{r}_i$.

\subsection{Objective Function}
\textbf{Overall objective function}. Given a sequence of $\numframes$ images, $\image_f$ ($1\leq \framenum \leq \numframes)$, we  optimize $\shapecoeff$ and the weights of the MLPs $\offsetmodel, \hairmodel, \texmodel, \defmodel$ jointly across the entire sequence, and $\posecoeff_{\framenum} \text{ and } \cam_{\framenum}$ per frame. We use the following overall objective function:
\begin{equation}
    \mathcal{L} = \mathcal{L}_{\text{recon}} + \mathcal{L}_{\text{ext}} + \mathcal{L}_{\text{int}} + \mathcal{L}_{\text{reg}},
\end{equation}
with reconstruction loss $\mathcal{L}_{\text{recon}}$, avatar exterior loss $\mathcal{L}_{\text{ext}}$, avatar interior loss $\mathcal{L}_{\text{int}}$ ($\mathcal{L}_{\text{int}}^{\text{body}}$ or $\mathcal{L}_{\text{int}}^{\text{face}}$) and regularization $\mathcal{L}_{\text{reg}}$.
For simplicity, we omit the frame index $\framenum$ and the optimization arguments whenever there is no ambiguity. For videos, the final objective function is the average over all frames. 

\vspace{0.75mm}
\noindent\textbf{Reconstruction loss}. We minimize the difference between the rendered image and the input image with the following objective:
\begin{equation}
    \mathcal{L}_{\text{recon}} = \lossweight{\text{pixel}}\cdot \mathcal{L}_{\delta}(\volrender-\image) + \lossweight{\text{semantic}} \cdot \mathcal{L}_{\text{semantic}}(\volrender, \image),
\end{equation}
where $\mathcal{L}_{\delta}$ is the Huber loss \cite{Huber1964} that penalizes the pixel-level difference. $\mathcal{L}_{\text{semantic}}$ is used to regularize the semantic difference. More specifically, we use an ID-MRF loss~\cite{Wang2018} $\mathcal{L}_{\text{mrf}}$ as $\mathcal{L}_{\text{semantic}}$ for reconstructing the body avatar, and an perceptual loss~\cite{johnson2016perceptual} $\mathcal{L}_{\text{per}}$ as $\mathcal{L}_{\text{semantic}}$ for reconstructing the head avatar.
While the Huber loss focuses on the overall reconstruction, the semantic loss allows us to reconstruct more details as previously shown by \citet{feng2021learning}.

\vspace{0.75mm}
\noindent\textbf{Avatar exterior loss}
Only minimizing the reconstruction error 
$\mathcal{L}_{\text{recon}}$ results in a NeRF that models the entire avatar including the body/face regions. 
Our goal is to only capture exterior components such as clothing or hair using $\hairmodel$. To achieve this, we employ a segmentation mask to explicitly limit the space within which the NeRF density can be. 
Given a segmentation mask $\imagemask_e$, which is represented by $\vect{1}$ for every exterior pixel (clothing or hair) and $\vect{0}$ elsewhere, we minimize the following exterior loss:
\begin{equation}
    L_{\text{ext}} = \lossweight{\text{ext}} \norm{\imagemask_v - \imagemask_e}_{1,1},
\end{equation}
with the rendered NeRF mask $S_v$, which is obtained by sampling rays for all image pixels and computing per ray 
\begin{equation}
\begin{aligned}
    &\bm{s_{v}}(\bm{r})=\sum_{k=1}^{n_b-1}T_k\big(1-\exp(-\sigma_k\Delta_k)\big).
\end{aligned}
\end{equation}
Minimizing $L_{\text{ext}}$ ensures that the aggregated density across rays (excluding the far bound) outside of clothing or hair is $0$. Therefore, only the intended exterior region is captured by the NeRF model.

\noindent\textbf{Avatar interior loss}. 
To further disentangle the avatar interior and exterior, we need to ensure that the interior mesh model does not capture any exterior variation. To this end, we define a few additional loss functions based on prior knowledge. 

First, the interior mesh should match the masked image. 
Given a binary mask $\imagemask$ of the entire avatar ($1$ for inside, $0$ elsewhere), we minimize the difference between the silhouette of the rendered body (denoted by $\meshrender^{s}(\vsmplx, \cam)$) and the given mask as
\begin{equation}
    \mathcal{L}_{\text{silhouette}} = \lossweight{\text{silhouette}} \mathcal{L}_{\delta}(\meshrender^{s}(\vsmplx, \cam) - \imagemask).
\end{equation} 
Second, the interior mesh should match visible avatar interior (\eg, for reconstructing the body, the body mesh should match the visible body region).
Only optimizing $\mathcal{L}_{\text{silhouette}}$ results in meshes that also fit the avatar exterior (\eg, clothing or hair). This is undesired especially for loose clothing or long hair, and also leads to visible artifacts when transferring clothing between subjects. 
Instead, given a binary mask $\imagemask_b$ of the visible body parts ($1$ for body parts, $0$ elsewhere), we minimize the following part-based silhouette loss
\begin{equation}\label{eq:intmask}
    \mathcal{L}_{\text{int-mask}} = \lossweight{\text{int-mask}} \mathcal{L}_{\delta}(\imagemask_b \odot \meshrender^{s}(\vsmplx, \cam) - \imagemask_b),
\end{equation}
and a part-based photometric loss
\begin{equation}\label{eq:skin}
    \mathcal{L}_{\text{skin}} = \lossweight{\text{skin}} \mathcal{L}_{\delta}(\imagemask_b \odot (\meshrender(\vsmplx, \col, \cam)-\image)),
\end{equation}
to put special emphasis on fitting visible interior parts.

Third, the interior mesh should stay within the exterior region. Specifically, the body or face should be generally covered by the clothing or hair, yielding to the following loss function:
\begin{equation}\label{eq:inside}
    \mathcal{L}_{\text{inside}} = \lossweight{\text{inside}} \mathcal{L}_{\delta}(ReLU(\meshrender^{s}(\vsmplx, \cam) - \imagemask_c)). 
\end{equation}
Fourth, the skin color of occluded body vertices should be similar to visible skin regions. 
For this, we minimize the difference between the body colors in occluded regions and the average skin color as
\begin{equation}
    \mathcal{L}_{\text{skin-inside}} = \lossweight{\text{skin-inside}} \mathcal{L}_{\delta}(\imagemask_c \odot (\meshrender(\vsmplx, \col, \cam)-\vect{C}_{\text{skin}})),
\end{equation}
where $\vect{C_{skin}}$
is the average color of the visible skin regions. 
In practice, we encountered challenges with skin detection not performing effectively. Therefore, for body video sequences, we assume that the hands are visible and utilize these hand regions to compute the average skin color. Moreover, for face videos, we determine the skin color by computing the mean color of the cheek region.

Combining the loss functions above, we use the following $\mathcal{L}_{\text{int}}$ for reconstructing the interior avatar:
\begin{equation}
    \mathcal{L}_{\text{int}} = \mathcal{L}_{\text{silhouette}} + \mathcal{L}_{\text{int-mask}} + \mathcal{L}_{\text{skin}} + \mathcal{L}_{\text{inside}} + \mathcal{L}_{\text{skin-inside}}.
\end{equation}

\vspace{0.75mm}

\noindent\textbf{Regularization}. We regularize the reconstructed mesh surface with
\begin{equation}
    \mathcal{L}_{\text{reg}} = \lossweight{\text{edge}} \mathcal{L}_{\text{edge}}(\vsmplx) 
                    + \lossweight{\text{offset}} \norm{\bm{\offsets}}_{2,2}, 
\end{equation}
where $\mathcal{L}_{\text{edge}}$ denotes the relative edge loss~\cite{Hirshberg2012_Coregistration} between the optimized interior mesh with and without the applied offsets. 
For the offset loss, we apply different weights to the body, hand and face region. Details are given in the experiment section.

\section{Intriguing Insights}

\noindent\textbf{Hybrid representation for general 3D modeling}. While the proposed \model demonstrates the effectiveness of hybrid 3D representation for human avatar modeling, the idea of hybrid representation can be broadly useful for modeling general 3D objects and scenes, especially for objects whose components have quite different physical properties. For example, a burning candle can be represented with a mesh-based candle and a NeRF-based flame, and a hourglass can be represented with mesh-based glass and point-based sand. \model shows the power of hybrid 3D representation through the lens of human avatar modeling, and we expect more future efforts can be put in exploring hybrid 3D representation.

\vspace{0.75mm}
\noindent\textbf{Hybrid vs. holistic 3D representation}. It has been a long-standing debate regarding the optimal holistic 3D representation for shape modeling. In the existing graphics pipeline, meshes are still a \emph{de facto} choice for holistic 3D representation due to its efficiency in storage and rendering. However, meshes can be quite limited in representing certain geometric structures, such as hair strand, fluid, smoke and complex clothing. Implicit 3D representations~\cite{park2019deepsdf,chen2019learning,mescheder2019occupancy,mildenhall2020nerf} demonstrate strong flexibility in complex shape representation, and in particular, NeRF further shows great novel view synthesis quality. However, it is difficult for NeRF to capture thin shell geometry like human body. While there is no single perfect 3D representation for all objects, why not combine the advantages of different representations and use them together? However, hybrid representation also inevitably introduces some shortcomings. First, the rendering process for hybrid representation becomes highly nontrivial and case-dependent. For example, our mesh-integrated volume rendering only works for the hybrid mesh and NeRF representation. Second, the representational heterogeneity makes subsequent learning and processing more difficult. For example, learning a generative model on hybrid representation is far more complicated than holistic representation. Moreover, editing hybrid representation will also become more challenging for designers. Third, how to choose the right 3D representations to combine is task-dependent. While \model uses meshes for human head and NeRFs for hair, it could be better to use a strand-based representation for hair.

\section{Experiments and Results} \label{sec:experiments}
\subsection{Datasets}
\label{sec:experimentsdatasetes} 
\model offers a solution for capturing dynamic objects from monocular video. We demonstrate the effectiveness of our approach by applying it to the challenging tasks of capturing clothing and hair from videos. To evaluate our approach, we introduce two types of datasets, one for full-body and one for head capture. 

\vspace{0.75mm}
\noindent\textbf{Full-body datasets}.
To compare with other state-of-the-art methods of realistic human capturing. We evaluate \model on sequences from public sources: People Snapshot~\cite{alldieck2018video}, iPER~\cite{lwb2019}, SelfRecon~\cite{jiang2022selfrecon}. However, none of them provide complicated clothes such as long dresses. Thus, we capture our own data MPIIS-SCARF, where we record videos of each subject wearing short and long dresses. 
For People Snapshot, we use the provided SMPL pose as initialization instead of running PIXIE~\cite{feng2021collaborative}. 
To be specific, we use 4 subjects (``male-3-casual'', ``female-3-casual'', ``male-4-casual'', ``female-4-casual'') from People Snapshot \cite{alldieck2018video} for qualitative and quantitative evaluation. The quantitative evaluation follows the settings of Anim-NeRF \cite{chen2021animatable}. 
We further use 4 subjects (``subject003'', ``subject016'', ``subject022'', ``subject023'') with outfit 1 and motion 1 from iPER \cite{lwb2019} and 4 synthetic video data (``female outfit1'', ``female outfit2'', ``female outfit3'', ``male outfit1'') and 1 self-captured video (``CHH female'') from SelfRecon \cite{jiang2022selfrecon} for qualitative evaluation.  
For MPIIS-SCARF, we use A-pose videos of subject ``Yao'' with six types of clothing for qualitative evaluation, those videos include loose dressing and short skirts. 
For each subject, we use around 100-150 images for optimization. 
For each frame, we run PIXIE~\cite{feng2021collaborative} to initialize $(\shapecoeff, \posecoeff, \expcoeff)$, and camera $\cam$.
For datasets without providing silhouette masks, we compute $\imagemask$ with \cite{RobustVideoMatting}, and \cite{clothsegmentation} for $\imagemask_c$. 

\vspace{0.75mm}
\noindent\textbf{Head datasets}.
We also evaluate \model on head videos from public sources. 
To be specific, we use video ``MVI\_1810" from IMAvatar~\cite{zheng2022avatar}, ``person\_0000'' and ``person\_0004'' from neural head avatar~\cite{grassal2022neural}.  As subjects with long hair are missing, we further collected one video with long hair from the Internet, named video ``b0\_0''~\cite{bili} (2:30). 
For each image from the video, we detect the upper body region and resize it to an image with 512x512 size. We then estimate 68 landmarks~\cite{bulat2017far} and iris~\cite{lugaresi2019mediapipe}, portrait matting with  MODNet~\cite{MODNet}, and segment face and hair with face parsing~\cite{faceparsing}.
Given the estimated labels and \smplx model, we roughly estimate the shape and texture parameters for the subject, and camera, pose, expression and lighting (Spherical harmonic) for each frame. Subsequently, for enhanced SMPL-X shape fitting, we perform parameter optimization across all frames, where shape and texture parameters are shared across frames. These optimized parameters serve as the initialization for our model training. 
Nonetheless, these videos often lack backviews of the head as they predominantly focus on face-related areas. To demonstrate our method's capacity for capturing complete hairs, we also incorporate synthetic data from the AGORA dataset~\cite{patel2020agora}. We select three subjects from Agora, each containing the mesh, texture, and corresponding SMPL fits. 200 images are rendered from the textured mesh for training \model. 

\subsection{Implementation Details}\label{sect:implementation}
We choose $\sigma=0.1$ and $|\mathcal{N}\left(\vect{x}\right)| = 6$. 
For full-body video, we set $t_n = -0.6$, and $t_f = 0.6$ and weight the individual losses with
$\lossweight{\text{pixel}}=1.0$,
$\lossweight{\text{semantic}}=0.0005$,
$\lossweight{\text{ext}}=0.5$,
$\lossweight{\text{silhouette}}=0.001$,
$\lossweight{\text{int-mask}}=30$,
$\lossweight{\text{skin}}=1.0$,
$\lossweight{\text{inside}}=40$,
$\lossweight{\text{skin-inside}}=0.01$,
$\lossweight{\text{edge}}=500$,
$\lossweight{\text{offset}}=400$. 
For $\lossweight{\text{offset}}$, the weight ratio of body, face and hands region is $2:3:12$. 
Note that it is important to perform the first stage NeRF training without optimizing the non-rigid deformation model. In this stage, we also set $\lossweight{\text{semantic}}=0$. 
In the second stage, the non-rigid deformation model then explains clothing deformations that cannot be explained by the body transformation. And $L_{semantic}$ helps capture more details that can not be modelled by the non-rigid deformation. 
The overall optimization time is around 40 hours with NVIDIA V100. 
In head video settings, we conducted SMPL-X fitting for all frames during data processing, that ensures accurate face fitting. By employing this as our initialization for DELTA training, we can directly train both mesh-based face and NeRF-based hair components. 
The chosen hyperparameters include $t_n = -1.5$, and $t_f = 1.5$. We assign weights to individual losses as follows:
$\lossweight{\text{pixel}}=1.0$,
$\lossweight{\text{semantic}}=0.015$,
$\lossweight{\text{ext}}=0.5$,
$\lossweight{\text{silhouette}}=0.001$,
$\lossweight{\text{int-mask}}=30$,
$\lossweight{\text{skin}}=1.0$,
$\lossweight{\text{inside}}=40$,
$\lossweight{\text{skin-inside}}=0.001$,
$\lossweight{\text{edge}}=500$,
$\lossweight{\text{offset}}=400$. 
To enhance training efficiency, we adopt Instant-NGP~\cite{mueller2022instant,li2023nerfacc} for parameterizing the hair component. Unlike the MLP layers in the original NeRF model, Instant-NGP leverages a hash table to store feature grids at various coarseness scales, resulting in fast training and inference speeds. We then require around 40 minutes of optimization time with NVIDIA A100. 

\begin{figure}[t]
    \includegraphics[width=\linewidth]{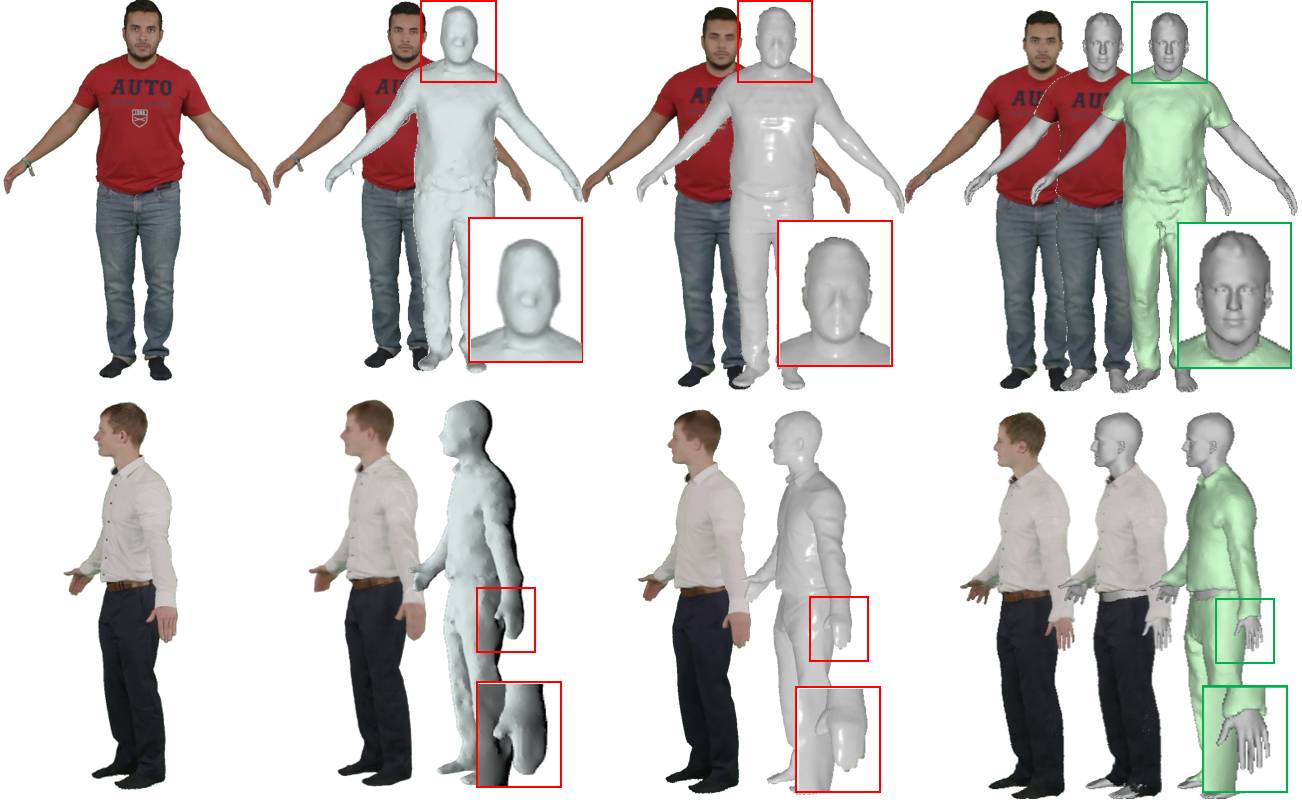}\\
    \raggedright
    \small{Reference image \hspace{1em} Anim-NeRF \hspace{2.5em} SelfRecon \hspace{4.5em} Ours }
    \vspace{-0.8em}
    \caption{Qualitative comparison with SelfRecon \cite{jiang2022selfrecon} and Anim-NeRF \cite{chen2021animatable} for reconstruction. While all methods capture the clothing with comparable quality, our approach has much more detailed face and hands due to the disentangled representation of clothing and body.}
    \vspace{-0.6em}
	\label{fig:comparison_selfrecon}
\end{figure}

\begin{table*}[t]
\small
\setlength{\tabcolsep}{2.5pt}
	\begin{center}
		\renewcommand{\arraystretch}{1.2}
		\begin{tabular}{l|ccccc|ccccc|ccccc}
			\multirow{2}{*}{Subject ID} & \multicolumn{5}{c|}  {PSNR$\uparrow$} & \multicolumn{5}{c|}{SSIM$\uparrow$} &
			\multicolumn{5}{c}{LIPIS$\downarrow$} \\
			                & NeRF  & SMPLpix   & NB    & Anim-NeRF    & {\cellcolor{Gray}}\model              & NeRF & SMPLpix & NB    & Anim-NeRF  & {\cellcolor{Gray}}\model           & NeRF & SMPLpix & NB    & Anim-NeRF  & {\cellcolor{Gray}}\model \\\shline
			male-3-casual   & 20.64 & 23.74     & 24.94 & 29.37     & {\cellcolor{Gray}}\textbf{30.59}    & .899 & .923  & .943 & .970   & {\cellcolor{Gray}}\textbf{.977} & .101 & .022  & .033 & \textbf{.017} & {\cellcolor{Gray}}.024 \\
			male-4-casual   & 20.29 & 22.43     & 24.71 & 28.37     & {\cellcolor{Gray}}\textbf{28.99}    & .880 & .910  & .947 & .961   & {\cellcolor{Gray}}\textbf{.970} & .145 & .031  & .042 & .027   & {\cellcolor{Gray}}\textbf{.025} \\
			female-3-casual & 17.43 & 22.33     & 23.87 & 28.91     & {\cellcolor{Gray}}\textbf{30.14}    & .861 & .929  & .950 & .974   & {\cellcolor{Gray}}\textbf{.977} & .170 & .027  & .035 & \textbf{.022}   & {\cellcolor{Gray}}.028 \\
			female-4-casual & 17.63 & 23.35     & 24.37 & 28.90     & {\cellcolor{Gray}}\textbf{29.96}    & .858 & .926  & .945 & .968   & {\cellcolor{Gray}}\textbf{.972} & .183 & .024  & .038 & \textbf{.017} & {\cellcolor{Gray}}.026 \\
		\end{tabular}
        \vspace{1mm}
		\caption{Quantitative comparison of novel view synthesis on People-Snapshot \cite{alldieck2018video}.}
		\label{tab:comparison_body}
	\end{center}
	\vspace{-4mm}
\end{table*}
\begin{figure}[t]
    \includegraphics[width=\linewidth]{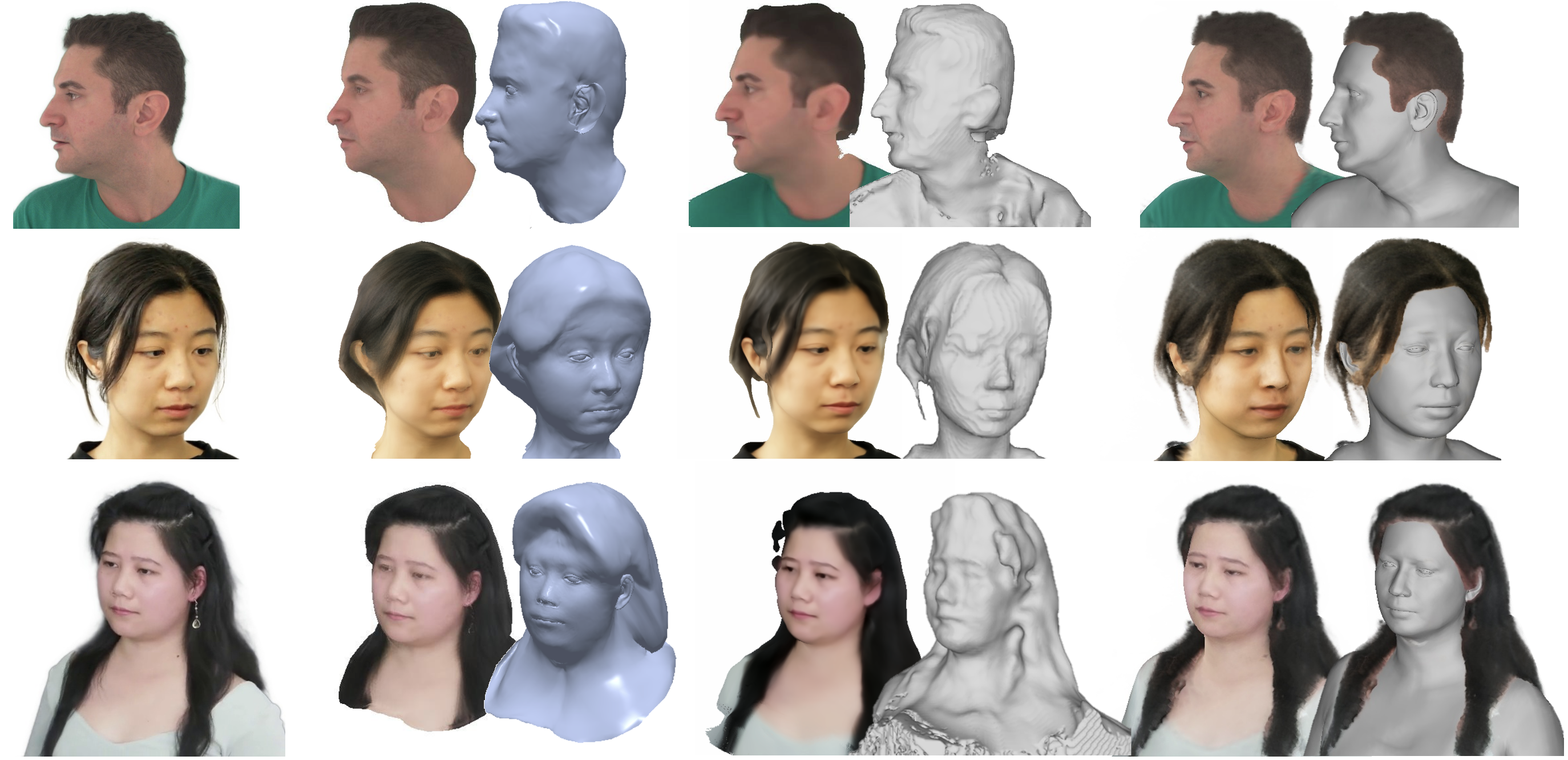}\\
    \raggedright
    \small{Reference image \hspace{2em} NHA
    \hspace{4em} IMAvatar \hspace{4em} Ours }
    \vspace{-0.8em}
    \caption{Qualitative comparison with neural head avatar (NHA)~\cite{grassal2022neural} and IMavatar~\cite{zheng2022avatar} for reconstruction. Our method exhibits superior performance in capturing the geometry of the face and shoulders. Moreover, it achieves exceptional rendering quality for the hair. This can be attributed to the effective utilization of a disentangled representation for separating the hair and face components in \model.}
    \vspace{-0.6em}
	\label{fig:comparison_face}
\end{figure}

\subsection{Comparison to Existing Methods} 

Our approach enables the creation of hybrid explicit-implicit avatars from monocular videos. We note that this has not been achieved by previous methods, which typically model clothed bodies or heads holistically using either implicit or explicit representations. To evaluate the effectiveness of our approach, we compare it to existing state-of-the-art methods on the challenging tasks of clothed-body and head modeling. 
The explicit-implicit modeling of \model also naturally disentangles objects such as the body and clothing, thereby enabling garment reconstruction. Unlike previous methods that reconstruct cloth geometry from a single image with the help of extensive 3D scan data, our approach can reconstruct garments from images alone. We evaluate the effectiveness of \model for garment reconstruction by comparing it to existing methods. 

\begin{figure}[t]
    \includegraphics[width=0.99\linewidth]{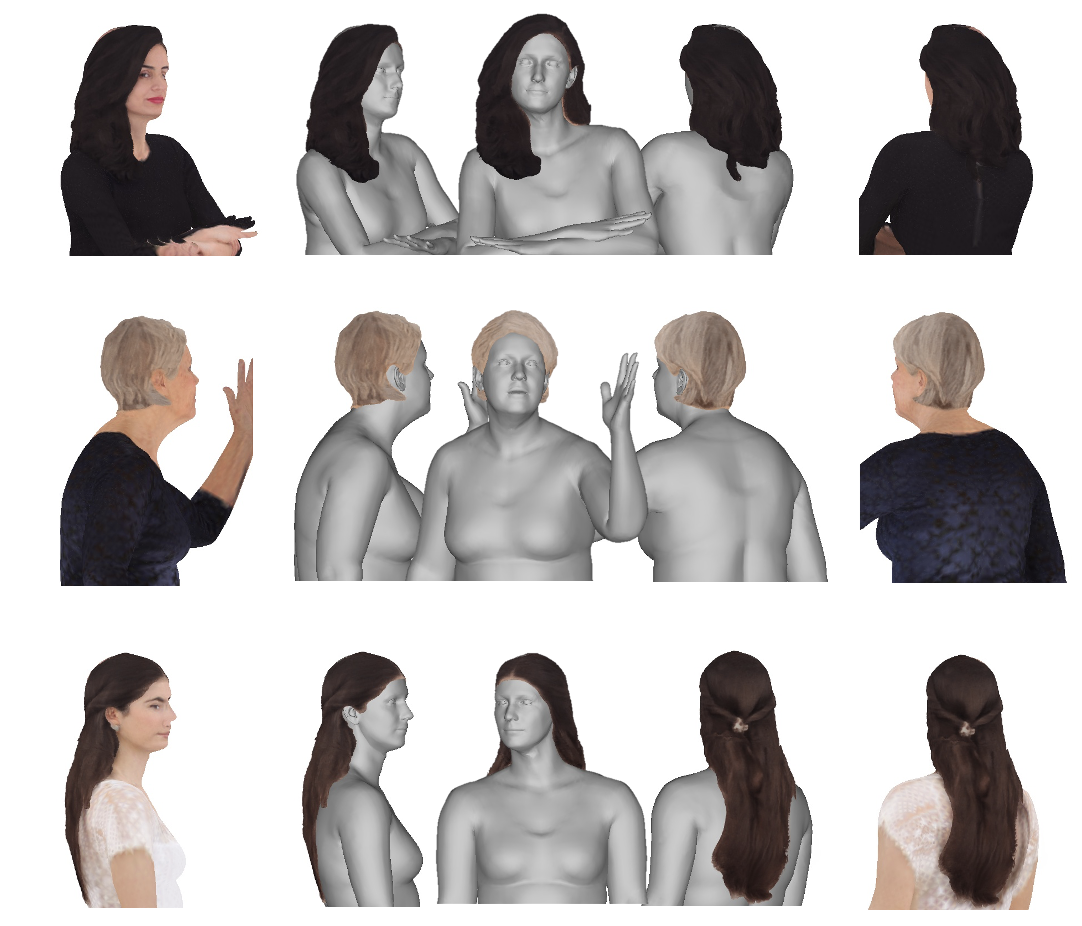}\\
    \vspace{-0.9em}
    \caption{Qualitative result on synthetic upper-body videos. The leftmost and rightmost images show the colored rendering of the learned avatars. The middle images show the hybrid rendering of the estimated upper body and hair. The results validate DELTA's ability to accurately represent complete hair views, including both short and long hair types. } 
    \vspace{-1.2em}
    \label{fig:qualitative_hair}
\end{figure}

\vspace{0.75mm}

\noindent\textbf{Body and clothing modeling}.
We quantitatively compare NB~\cite{omran2018neural}, SMPLpix~\cite{SMPLpix:WACV:2020}, Neural Body~\cite{peng2021neural} and Anim-NeRF~\cite{chen2021animatable}, following the evaluation protocol of~\cite{chen2021animatable}. To be specfic, we use 4 subjects (``subject003'', ``subject016'', ``subject022'', ``subject023'') with outfit 1 and motion 1 from iPER \cite{lwb2019} for qualitative evaluation.  
For all subjects, we uniformly select frames 1-490 with a step-size 4 for optimization. We use 4 synthetic video data (``female outfit1'', ``female outfit2'', ``female outfit3'', ``male outfit1'') and 1 self-captured video (``CHH female'') from SelfRecon \cite{jiang2022selfrecon}. For each subject, we use 100 frames for optimization. For self-captured data, we use A-pose videos of subject ``Yao'' with six types of clothing for qualitative evaluation, those videos include loose dressing and short skirts. For each video, we uniformly select frames 0-400 with a step-size 2 for optimization. 
Table~\ref{tab:comparison_body} shows that \model is more accurate than the other methods under most metrics.
The qualitative comparison in Figure~\ref{fig:comparison_selfrecon} demonstrates that \model can better reconstruct the hand and face geometry compared to SelfRecon~\cite{jiang2022selfrecon} and Anim-NeRF~\cite{chen2021animatable}.

\vspace{0.75mm}
\noindent\textbf{Face and hair modeling}.
We conduct an evaluation of our proposed method using four real-world videos. To assess the effectiveness of our approach, we compare it with two state-of-the-art methods,  neural head avatar~(NHA)~\cite{grassal2022neural} and IMavatar~\cite{zheng2022avatar}. 
To ensure a fair comparison, we adopt the same experimental protocol, where we train NHA and IMavatar using exactly the same set of video frames and reserve the remaining frames for evaluation.
To be specific, for subjects ``person\_0000'', ``person\_0004'' and ``MVI\_1810'', we sample every 50 frames for evaluation, and for the subject ``b0\_0", we sample every 5 frames. 
Following neural head avatar~\cite{grassal2022neural}, for each image, we keep the trained model and optimize per-frame parameters such as camera, pose, and expression. 
Consistent with prior research~\cite{gafni2021dynamic, zheng2022avatar, grassal2022neural}, we employ four image-based metrics to evaluate our approach. These metrics include pixel-wise L1 loss, peak signal-to-noise ratio (PSNR), structural similarity metric (SSIM), and the learned perceptual image patch similarity (LPIPS).
We find that NHA only focuses on the face, neck, and hair regions for training and evaluation. For a fair comparison, we compute the metrics on both the whole human region and only face, neck and hair regions. 
\begin{table*}
\small
\centering
\setlength{\tabcolsep}{9.25pt}
\renewcommand{\arraystretch}{1.2}
\begin{tabular}{l|l|cccc|cccccccc}
\multirow{2}{*}{Video}  & \multirow{2}{*}{Model} & \multicolumn{4}{c|}{Whole} & \multicolumn{4}{c}{Face, Hair and Neck} \\ 
  & & 
  {L1~$\downarrow$} & {PSNR~$\uparrow$} & {SSIM~$\uparrow$}  & {LIPIS~$\downarrow$} &
  {L1~$\downarrow$} & {PSNR~$\uparrow$} & {SSIM~$\uparrow$}  & {LIPIS~$\downarrow$} \\\shline
\multirow{3}{*}{person\_0000} 
& NHA~\cite{grassal2022neural}
& 0.094 & 12.15 & 0.843 & 0.198
& \textbf{0.012} & \textbf{24.92} & \textbf{0.920} & \textbf{0.046}
\\
& IMavatar~\cite{zheng2022avatar} 
& 0.024 & 22.55 & 0.882 & 0.177
& 0.015 & 23.70 & 0.917 & 0.089
\\
&{\cellcolor{Gray}}\model
& {\cellcolor{Gray}} \textbf{0.021} & {\cellcolor{Gray}} \textbf{24.04} & {\cellcolor{Gray}} \textbf{0.892}
& {\cellcolor{Gray}} \textbf{0.122} 
& {\cellcolor{Gray}} 0.017 & {\cellcolor{Gray}} 23.37
& {\cellcolor{Gray}} 0.914 & {\cellcolor{Gray}} 0.086 
\\

\hline

\multirow{3}{*}{MVI\_1810} 
& NHA~\cite{grassal2022neural} 
& 0.054 & 16.01 & 0.817 & 0.195
& 0.038 & 18.94 & 0.842 & 0.149
\\
& IMavatar~\cite{zheng2022avatar} 
& \textbf{0.039} & 20.33 & 0.829 & 0.171
& \textbf{0.031} & 21.44 & 0.851 & 0.137
\\
& {\cellcolor{Gray}}\model
& {\cellcolor{Gray}} \textbf{0.039} & {\cellcolor{Gray}} \textbf{21.33} & {\cellcolor{Gray}} \textbf{0.835} & {\cellcolor{Gray}} \textbf{0.156} 
& {\cellcolor{Gray}} 0.034 & {\cellcolor{Gray}} \textbf{22.12}
& {\cellcolor{Gray}} \textbf{0.852} & {\cellcolor{Gray}} \textbf{0.132} 
\\
\hline

\multirow{3}{*}{b0\_0} 
& NHA~\cite{grassal2022neural} 
& 0.062 & 15.60 & 0.874 & 0.203 
& 0.042 & 16.12 & 0.896 & 0.137
\\
& IMavatar~\cite{zheng2022avatar} 
& 0.043 & 19.61 & 0.871 & 0.188 
& 0.030 & 20.13 & 0.905 & \textbf{0.097}
\\
& {\cellcolor{Gray}}\model
& {\cellcolor{Gray}} \textbf{0.025} & {\cellcolor{Gray}} \textbf{23.28} & {\cellcolor{Gray}} \textbf{0.909}  & {\cellcolor{Gray}} \textbf{0.096} 
& {\cellcolor{Gray}} \textbf{0.022} & {\cellcolor{Gray}} \textbf{21.47} & {\cellcolor{Gray}}\textbf{0.917}  & {\cellcolor{Gray}} 0.103
\end{tabular}
\vspace{1mm}
\caption{Quantitative comparison of novel pose and expression synthesis on public real videos.}
\vspace{-3mm}
\label{tab:comparison_quan_face}
\end{table*}

\begin{figure}[t]
    \centering    \includegraphics[width=0.98\columnwidth]{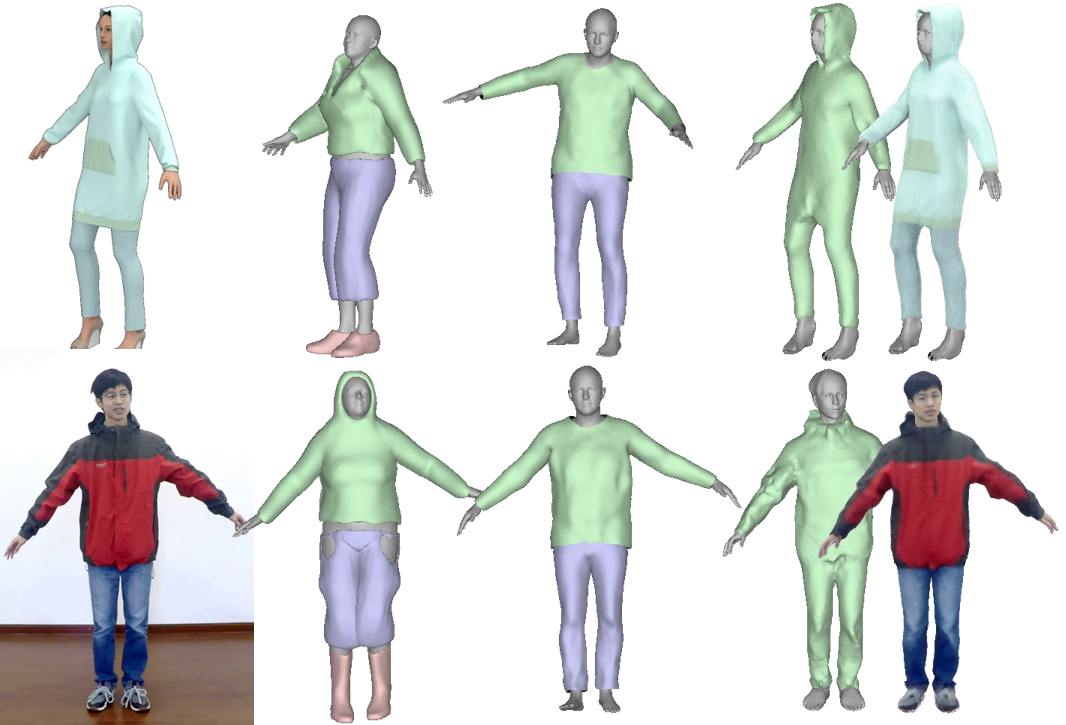}\\
    \small{Input image \hspace{1.5em} SMPLicit \hspace{3em} BCNet \hspace{5em} Ours \hspace{2em}}
    \vspace{-0.65em}
	\caption{Qualitative comparison of garment reconstruction. \model reconstructs different clothing types more faithfully than SMPLicit \cite{corona2021smplicit} and BCNet \cite{jiang2020bcnet}.}
    \vspace{-0.45em}
	\label{fig:comparison_garment}
\end{figure}

The quantitative comparison presented in Table \ref{tab:comparison_quan_face} demonstrates that our method attains the highest level of quality when considering the entire human region. However, when specifically focusing on the face, hair, and neck regions, it is worth noting that NHA achieves superior results for subjects with short hair, such as ``person\_0000''. 
Nevertheless, when it comes to subjects with longer hair, NHA struggles to capture both hair and face details, as exemplified in instances such as ``MVI\_1810'' and ``b0\_0''. 
In contrast, our method performs effectively across various hair types and successfully captures the entirety of the avatar, including changes in the shoulders. This capability can be attributed to the utilization of hybrid representations within our approach.

We additionally provide qualitative comparisons for novel view images and shapes in Figure \ref{fig:comparison_face}, along with supplementary qualitative results of DELTA applied to synthetic upper-body videos from the AGORA~\cite{patel2020agora} dataset in Figure \ref{fig:qualitative_hair}. Our method showcases superior performance in capturing accurate face and shoulder geometry, while also delivering high-quality renderings of the hair. 

\subsection{Applications}
\begin{figure}[t]
    \includegraphics[width=\linewidth]{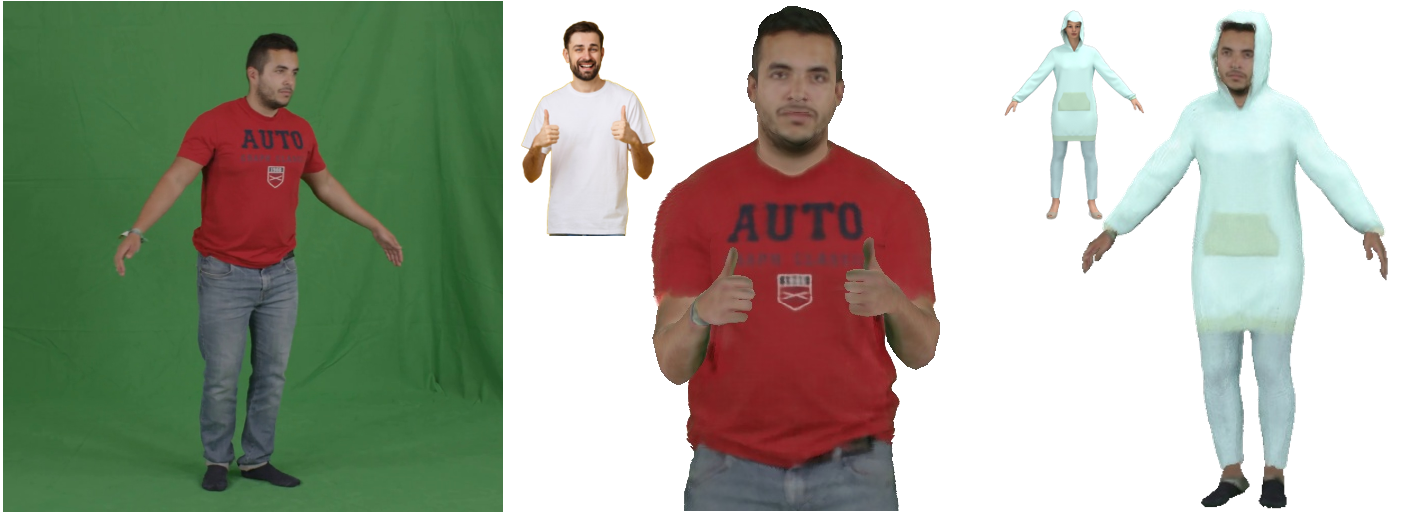}\\
    \small{Source subject \hspace{5em} Reposing \hspace{4em} Clothing transfer}
    \vspace{-0.8em}
    \caption{Applications of DELTA. The hybrid representation enables (middle) reposing with detailed control over the body pose and (right) dressing up the source subject with target clothing. The target pose and clothing are shown in the inset images.}
    \vspace{-0.75em}
    \label{fig:applications}
\end{figure}

\begin{figure}[t]
    \includegraphics[width=\linewidth]{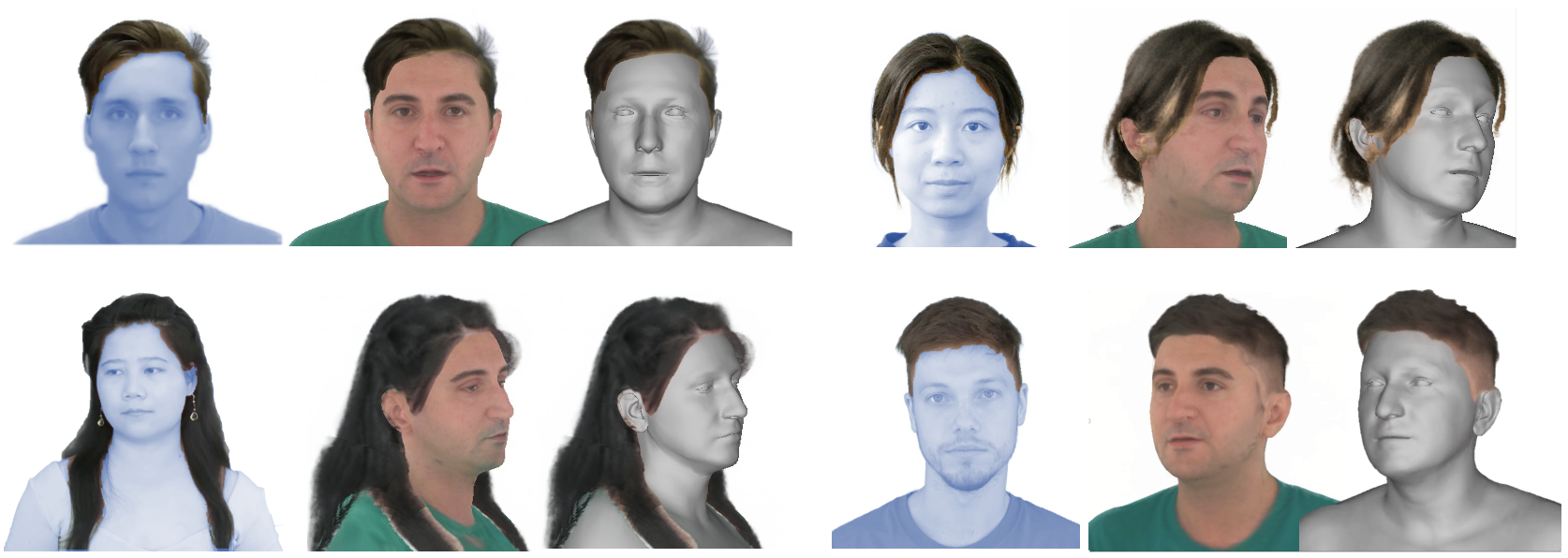}\\
    \vspace{-0.8em}
    \caption{Applications of DELTA. The hybrid representation enables transferring NeRF-based hairs into another face. Picture in the left indicates the source of the original hair. The avatar can also be animated with different poses and expressions. }
    \vspace{-0.5em}
    \label{fig:hair_transfer}
\end{figure}

\vspace{0.75mm}
\noindent\textbf{Body and garment reconstruction}.
We show comparisons on Garment reconstruction with SMPLicit \cite{corona2021smplicit} and BCNet \cite{jiang2020bcnet} in Fig~\ref{fig:comparison_garment}.
\model gives better visual quality than SMPLicit and BCNet. 
Note that the training/optimization settings are different, they reconstruct the body and garment from a single image, while our results are learned from video. 
However, they require a large set of 3D scans and manually designed cloth templates for training, while we do not need any 3D supervision, and capture the garment appearance as well. 
Figure~\ref{fig:comparison_garment} shows that \model reconstructs different clothing types more faithfully.

\vspace{0.75mm}

\noindent\textbf{Reposing}.
For clothed body modeling, unlike previous methods that represent clothed bodies holistically, \model offers more fine-grained control over body pose especially hand pose. 
Figure~\ref{fig:applications} shows reposing into novel poses.
Similar to the face and hair, utilizing an explicit shape model to present face region facilitates generalization across a wide range of facial expression animations. As Figure~\ref{fig:hair_transfer} shows different expressions of the reconstructed avatar. 

\begin{figure}[t]
    \includegraphics[width=\linewidth]{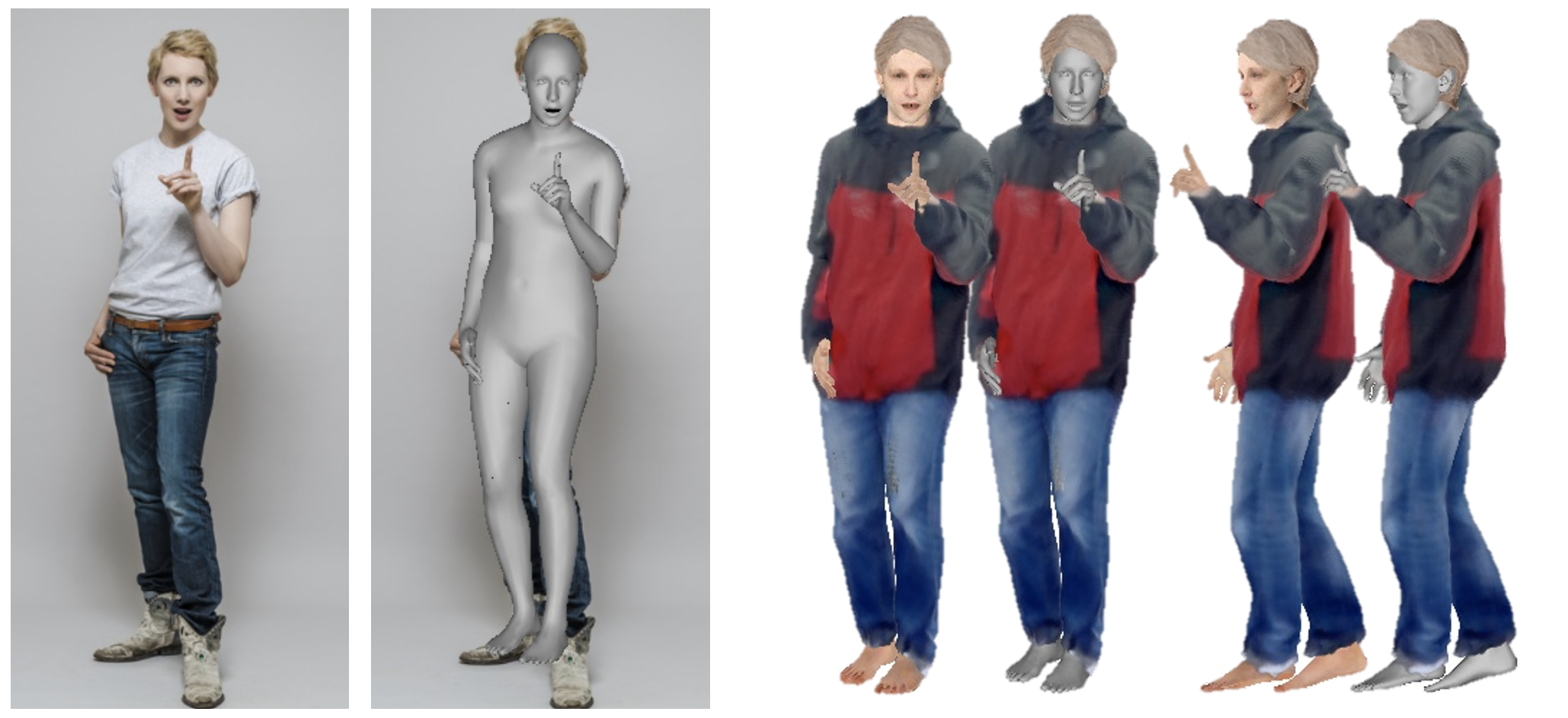}\\
    \vspace{-0.6em}
    \caption{Virtual try-on Application of DELTA. Given a single image, we can estimate the body shape using PIXIE~\cite{feng2021collaborative}. The body texture is from PIXIE template. Both the trained hair and clothing can be subsequently applied to this subject, resulting in smooth virtual try-on applications. In this instance, the captured hair is derived from the second example in Figure \ref{fig:qualitative_hair}, and the clothing is from the second example of Figure \ref{fig:comparison_garment}.
    }
    \vspace{-0.5em}
    \label{fig:both_transfer}
\end{figure}

\vspace{0.75mm}
\noindent\textbf{Clothing and hair transfer}.
Figures \ref{fig:teaser}, \ref{fig:applications} and \ref{fig:hair_transfer} qualitatively demonstrate the capability of our hybrid 3D representation in enabling clothing and hair transfer between avatars. We note that the clothing and hair is able to seamlessly adapt to accommodate various body shapes. 
Furthermore, the trained hair and clothing models can be both seamlessly transferred to different subjects. One potential application involves utilizing an existing body estimation method like PIXIE~\cite{feng2021collaborative} to estimate the body shape from a single image. Subsequently, our captured hair and clothing models can be applied to this subject, offering a streamlined approach for virtual try-on applications, as shown in Figure~\ref{fig:both_transfer}.

\begin{figure}[t]
    \includegraphics[width=\linewidth]{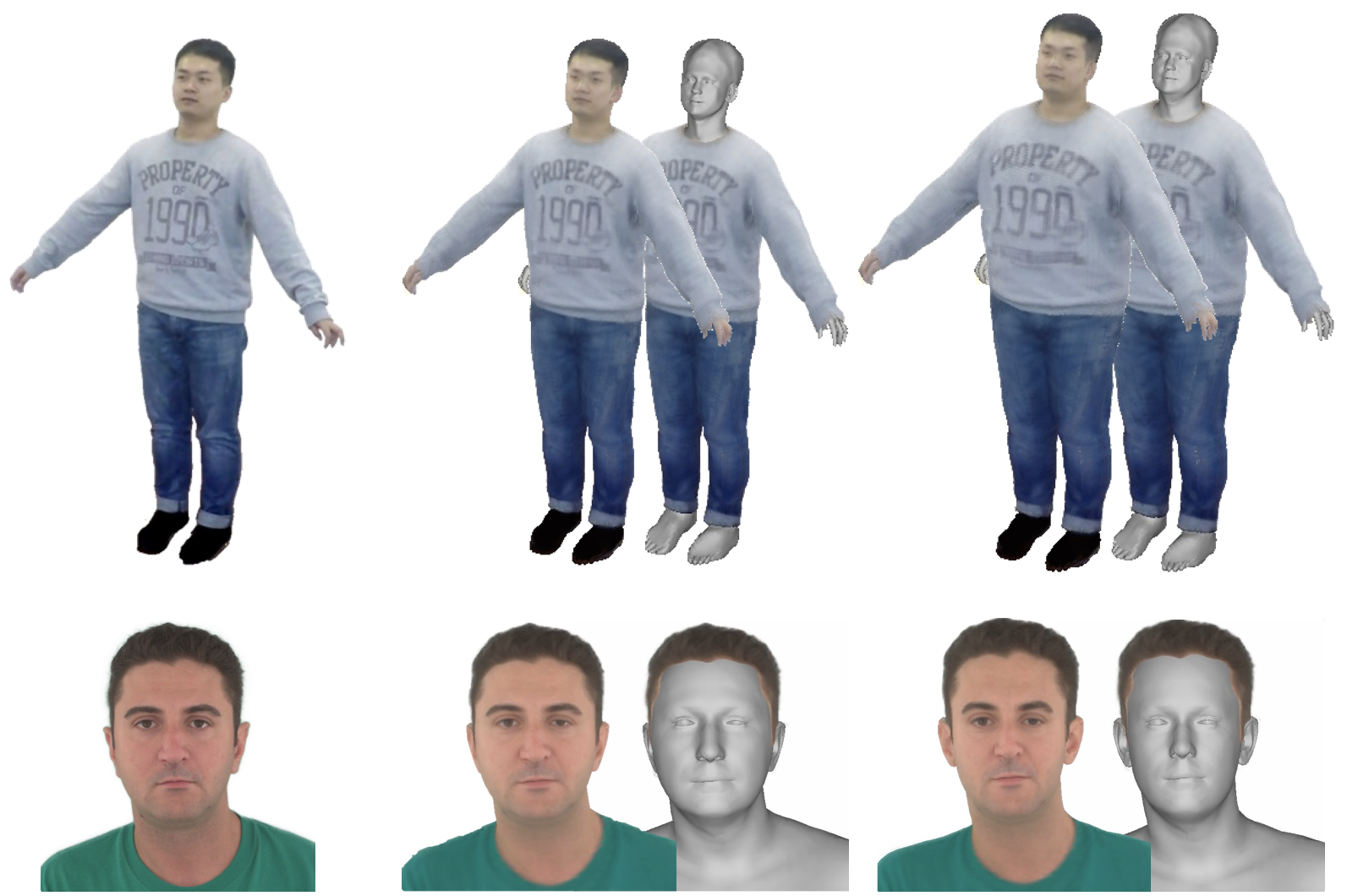}\\
    \small{Source subject \hspace{3.5em} Reconstruction \hspace{4em} Changing shape}
    \vspace{-0.7em}
    \caption{\model can change underlying body/face shapes by modifying SMPL-X shape parameters, and the NeRF-based clothing/hair will adapt to the new body/face shape accordingly.}
    \vspace{-0.5em}
    \label{fig:alter_shape}
\end{figure}

\vspace{0.75mm}
\noindent\textbf{Altering human Shape}.
Figure~\ref{fig:alter_shape} highlights an additional facet of DELTA's capabilities. We show the capacity to alter human body or face shape through adjustments in SMPL-X shape parameters. Subsequently, the NeRF-based clothing or hair seamlessly adjusts to align with the modified shape.

\subsection{Ablation Study}
We run different ablation experiments to show the impact of different components of our hybrid representation, and to show the impact of the pose refinements. 

\begin{figure}[t]
    \includegraphics[width=\columnwidth]{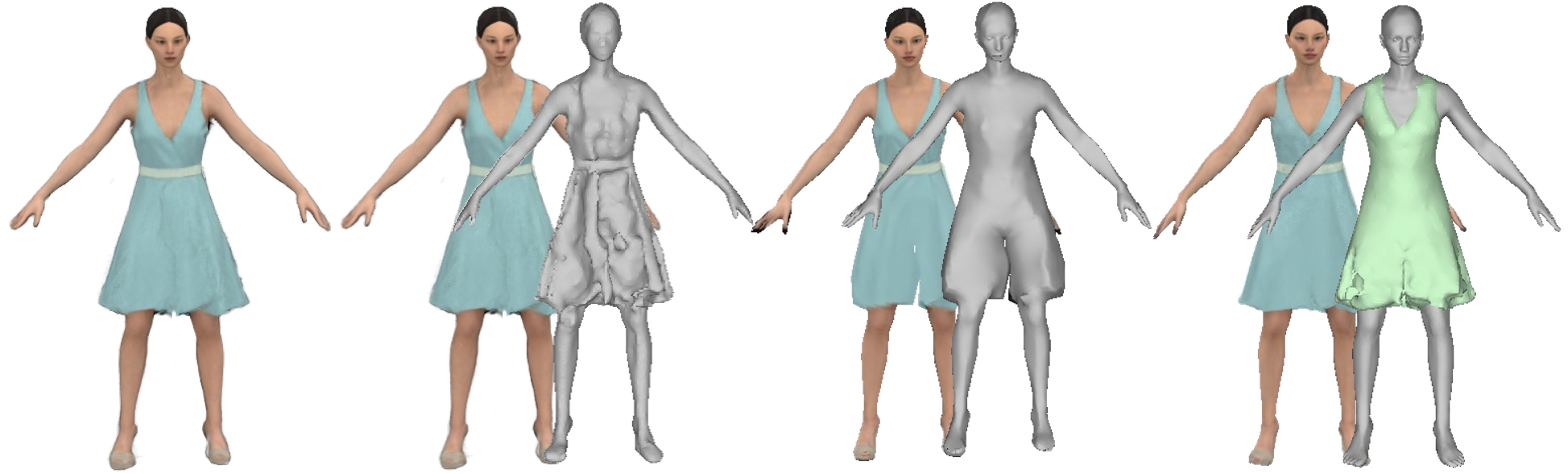}\\
    \raggedright
    \small{Reference image \hspace{2.5em} NeRF \hspace{3.3em} SMPL-X +D \hspace{4.2em} Ours}
     \vspace{-0.05in}
    \caption{Rendered images and extracted meshes from different components of \model.
    Our hybrid representation gives a better estimated face, hand, and clothing geometry than vanilla NeRF or a mesh-based representation. }
    \label{fig:ablation_representation}
 	\vspace{-0.8em}
\end{figure}

\vspace{0.75mm}
\noindent\textbf{Effect of representations}.
\model consists of a NeRF to represent clothing, and a mesh with vertex displacements. 
Figure~\ref{fig:ablation_representation} compares NeRF to holistically represent body and clothing (i.e., \model w/o body-clothing segmentation) and mesh-only based representation (i.e., \model w/o NeRF). 
Our hybrid representation is better able to estimate the face, hands, and complex clothing. 
Note that, unlike our hybrid representation, none of the existing body NeRF methods can transfer clothing between avatars.

\begin{figure}[t]
    \includegraphics[width=\linewidth]{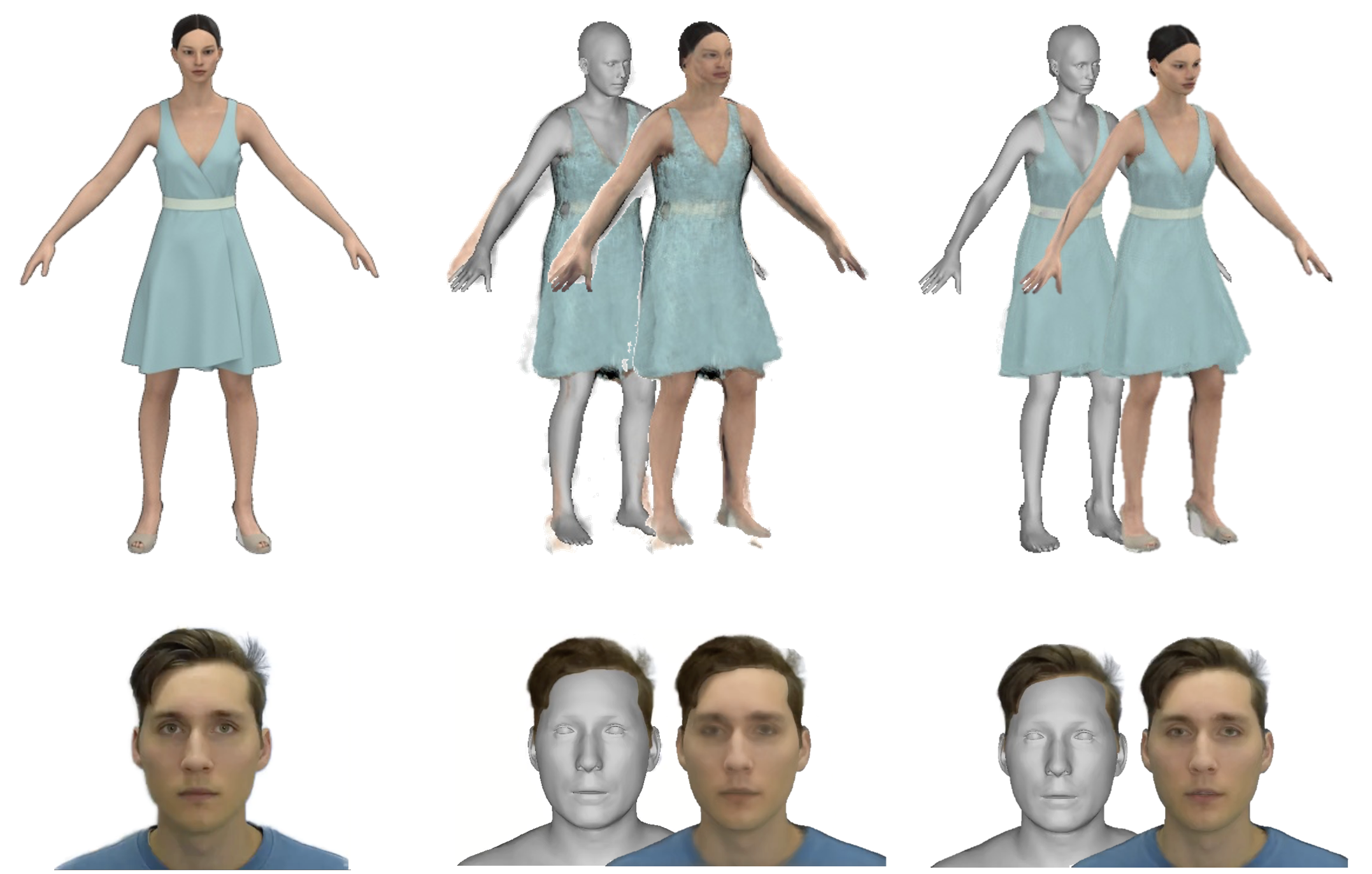}\\
    \small{Source subject \hspace{1.8em} w/o pose refinement \hspace{2.0em} w/ pose refinement}
    \vspace{-0.8em}
	\caption{Rendering results of clothed body (up) and head (bottom) w/o and w/ pose refinement. The pose refinement improves the visual quality of the reconstruction, as more texture details are reconstructed. For the face subject, please zoom in to check the difference.}
	\label{fig:ablation_pose}
	\vspace{-0.6em}
\end{figure}

\vspace{0.75mm}
\noindent\textbf{Effect of pose refinement}.
Since the pose estimation for each frame is not accurate, the pose refinement is important to gain details. 
We try learning our method without pose refinement. 
Figure~\ref{fig:ablation_pose} shows that pose refinement improves the image quality a lot.

\section{Discussion and Limitation}

\begin{figure}[t]
    \includegraphics[width=\linewidth]{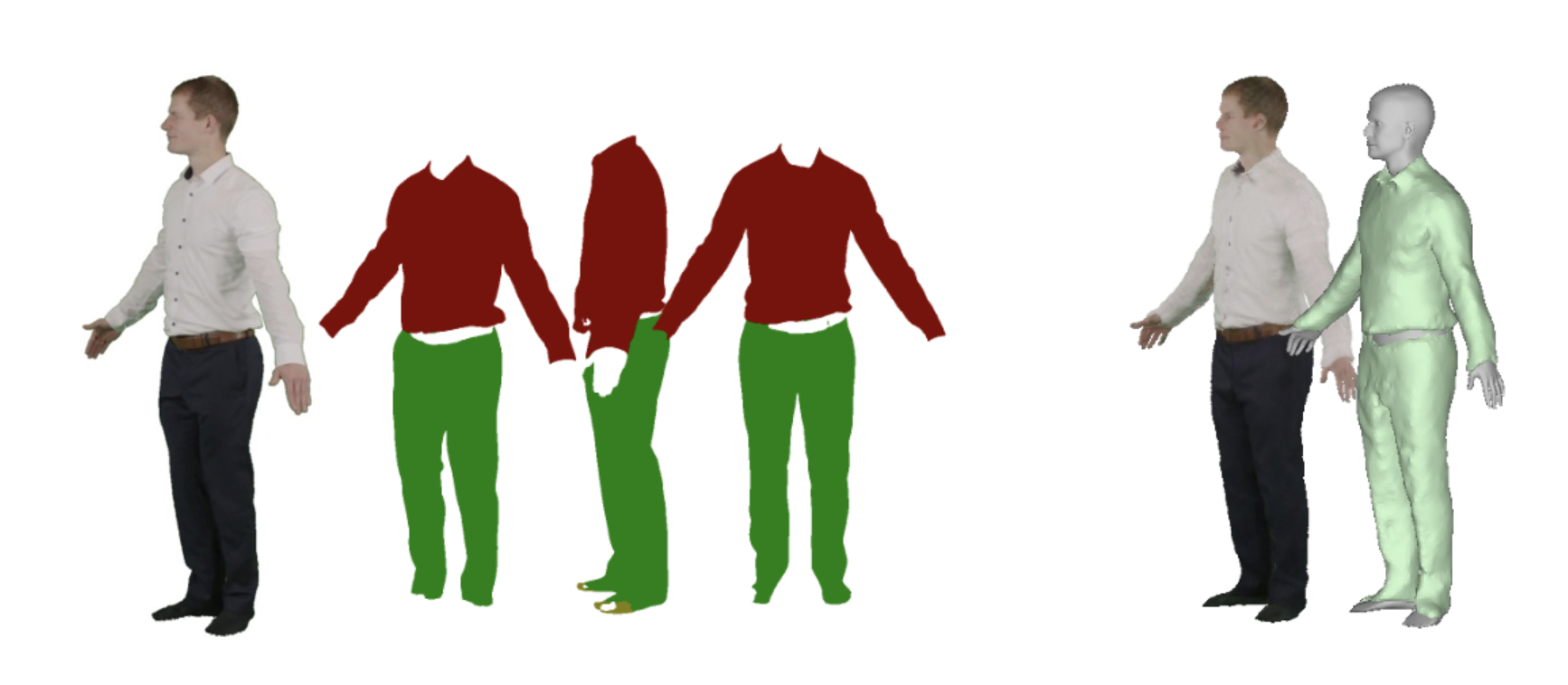}\\
    \small{\hspace{1em}Input image and clothing segmentations \hspace{4em} Reconstruction}
    \vspace{-0.8em}
	\caption{The wrong clothing segmentation masks result in a visible gap within the
reconstructed clothing.}
	\label{fig:limitation_segmentation}
	\vspace{-0.6em}
\end{figure}

\noindent\textbf{Segmentation}.
\model requires body and clothing/hair segmentation for training. 
Segmentation errors of the subject and background negatively impact the visual quality of the extracted avatar, and erroneous clothing or hair segmentation results in poor separation of mesh-based body and NeRF-based clothing or hair part. Figure~\ref{fig:limitation_segmentation} shows the wrong reconstruction due to consistent clothing segmentation errors, e.g. the belt is not recognized as part of clothing in segmentation, this results in wrong disentanglement between human body and clothing. 
Enforcing temporal consistency by exploiting optical flow could improve the segmentation quality.

\begin{figure}[t]
    \includegraphics[width=\linewidth]{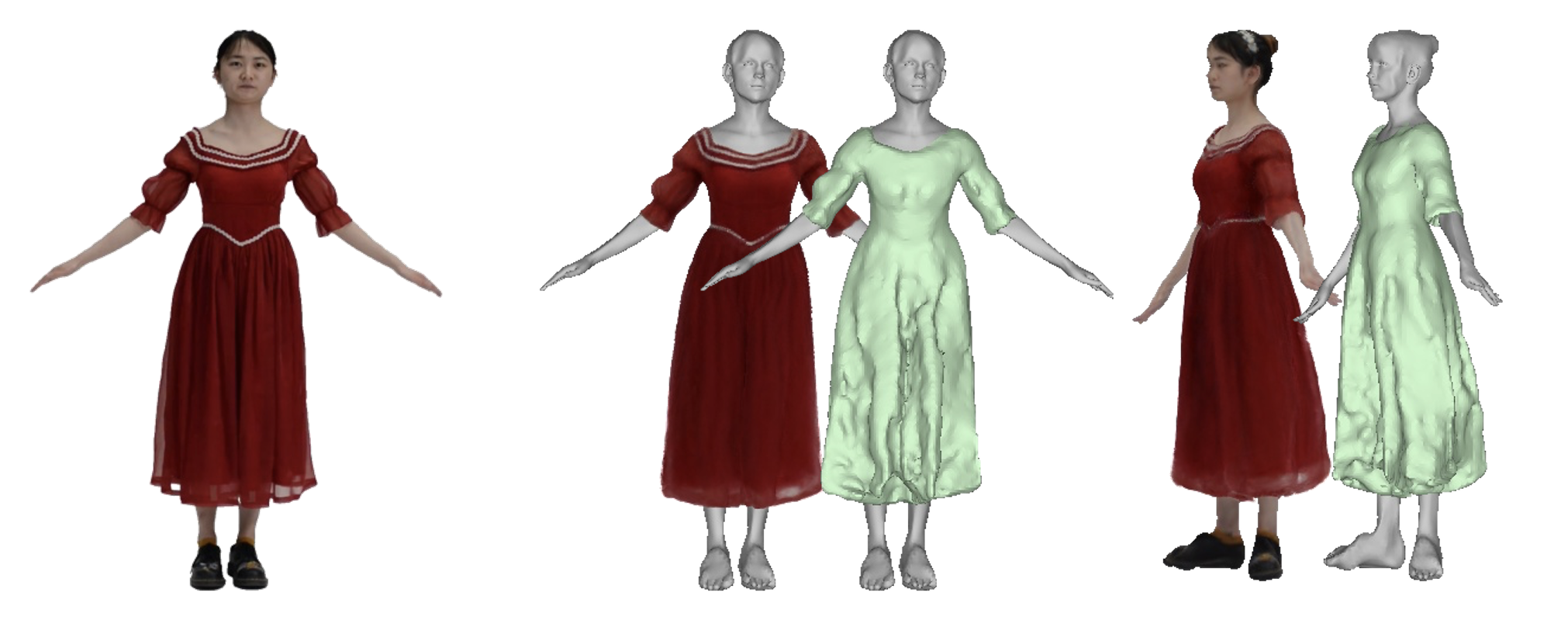}\\
    \small{Reference image \hspace{4em} Captured clothing appearance \& geometry \hspace{5em} }
    \vspace{-1.5em}
	\caption{While \model gives good visual quality for clothing renderings, the underlying geometry of the NeRF clothing is sometimes noisy.}
	\label{fig:limitation_geometry_clothing}
	\vspace{-0.4em}
\end{figure}

\begin{figure}[t]
    \includegraphics[width=\linewidth]{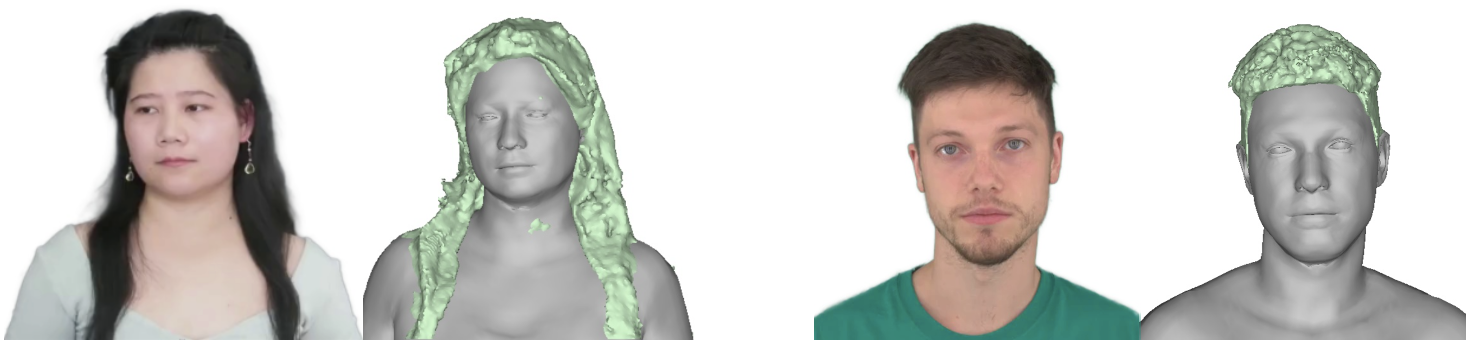}\\
    \vspace{-0.8em}
	\caption{Two examples of captured hair appearance and geometry. While \model gives good visual quality for hair renderings, the underlying geometry of the NeRF hair is  noisy.}
	\label{fig:limitation_geometry_hair}
	\vspace{-0.6em}
\end{figure}

\vspace{0.75mm}
\noindent\textbf{Geometric quality}.
The strength of NeRF is its visual quality and the ability to synthesize realistic images, even when the geometry is not perfect. 
Figure~\ref{fig:limitation_geometry_clothing} and Figure~\ref{fig:limitation_geometry_hair} show examples of noisy geometry despite good visual quality.  
In contrast, recent SDF-based methods have demonstrated good geometric reconstruction (e.g., \cite{jiang2022selfrecon}). 
It may be possible to leverage their results to better represent the underlying clothed shape or to regularize NeRF. 


\begin{figure}[t]
    \includegraphics[width=\linewidth]{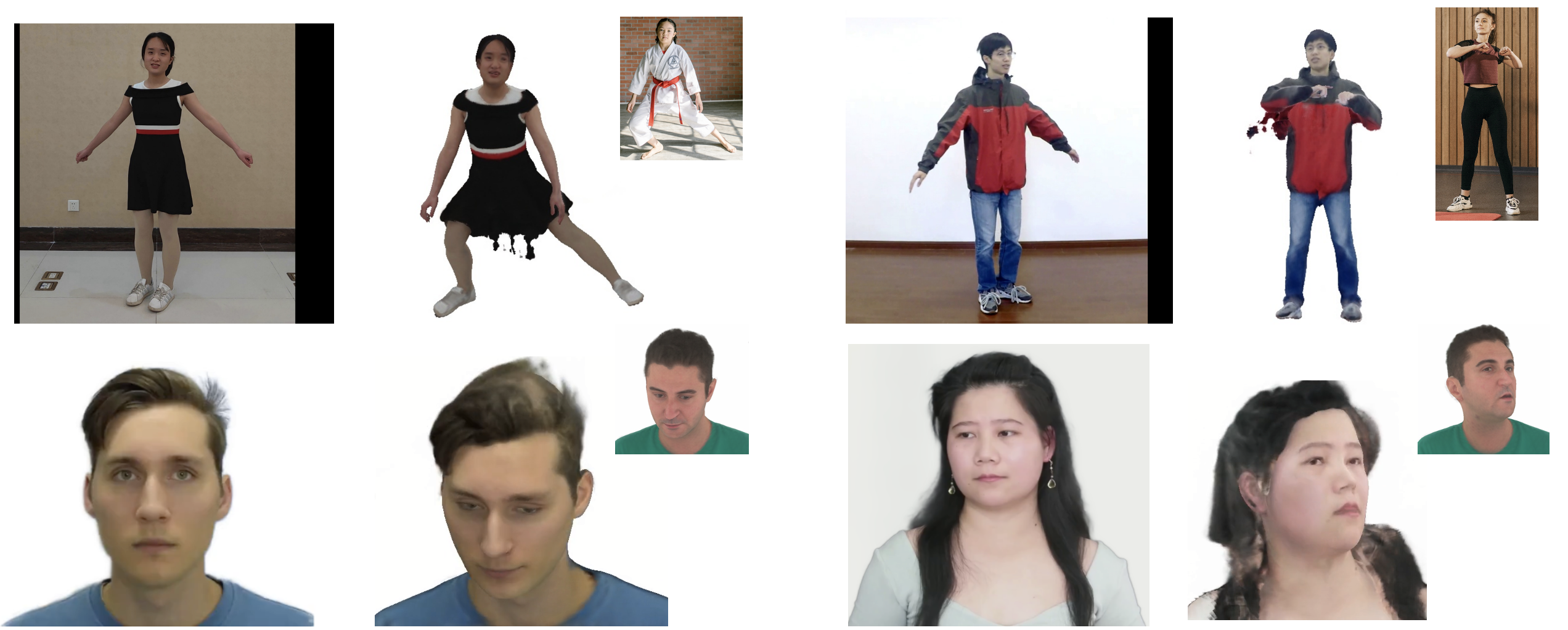}\\
    \vspace{-0.8em}
	\caption{Some failure cases for avatar animation. Artifacts may occur in extreme poses. In the example shown at the bottom-left, the top part of the hair is absent since it was never observed in the training data.}
	\label{fig:limitation_animation}
	\vspace{-0.6em}
\end{figure}

\vspace{0.75mm}
\noindent\textbf{Novel poses and views}.
Although \model demonstrates generalization to unseen poses, artifacts may occur in extreme poses. As depicted in Figure \ref{fig:limitation_animation}, the animation results for new poses exhibit satisfactory performance for the body and face regions. However, artifacts are prevalent in the non-rigid fusion (clothing or hair) component. Notably, for regions that have not been encountered in the training data, our model will fail to capture the desired details. For instance, in the example featuring short hair, the hair in the head top is always missing in all poses and views in the video. To address these limitations, potential solutions include incorporating regularization techniques during NeRF optimization or training a generative model using a diverse set of training examples encompassing different individuals and poses. These approaches have the potential to enhance the robustness and accuracy of the model when dealing with unseen regions and extreme poses. 

\vspace{0.75mm}
\noindent\textbf{Pose initialization}.
\model refines the body pose during optimization. However, it may fail if the initial pose is far from the right pose. 
Handling difficult poses where PIXIE~\cite{feng2021collaborative} fails requires a more robust 3D body pose estimator. 

\vspace{0.75mm}
\noindent\textbf{Dynamics}.
\model handles non-rigid cloth deformation with the pose-conditioned deformation model. 
While the global pose can account for some deformation, how to accurately model the clothing and hair dynamics as a function of body movement remains an open problem and is an important future work.

\vspace{0.75mm}
\noindent\textbf{Lighting}.
As with other NeRF methods, we do not factor lighting and material properties.
This results in baked-in shading and the averaging of specular reflections across frames.
Factoring lighting from shape and material is a key next step to improve realism. 

\vspace{0.75mm}
\noindent\textbf{Facial expressions}.
\model uses the facial expressions estimated by PIXIE \cite{feng2021collaborative} which is unable to capture the full spectrum of emotions (cf.~\cite{danvevcek2022emoca}).
Also, we have not fully exploited neural radiance fields to capture complex changes in facial appearance, \eg, due to the movement of mouth opening. We believe this is a promising future direction.

\section{Concluding Remarks}
\model is able to automatically extract human body, clothing or hair from a monocular video. 
Our key novelty is a hybrid representation that combines a mesh-based body model with a neural radiance field to separately model the body and clothing/hair. 
This factored representation enables \model to transfer clothing/hair between avatars, animate the body pose of the avatars including finger articulation, alter their body shape and facial expression, and visualize them from unseen viewing directions. 
This property makes \model well suited to VR and virtual try-on applications.
Finally, \model outperforms existing avatar extraction methods from videos in terms of visual quality and generality.

\begin{acks}
We would like to sincerely thank Sergey Prokudin, Yuliang Xiu, Songyou Peng, Qianli Ma for fruitful discussions, and Peter Kulits, Zhen Liu, Yandong Wen, Hongwei Yi, Xu Chen, Soubhik Sanyal, Omri Ben-Dov, Shashank Tripathi for proofreading. 
We also thank Betty Mohler, 
Sarah Danes, Natalia Marciniak, Tsvetelina Alexiadis,
Claudia Gallatz, and Andres Camilo Mendoza Patino for their supports with data.
This work was partially supported by the Max Planck ETH Center for Learning Systems.

\vspace{0.75mm}
\noindent\textbf{Disclosure}. MJB has received research gift funds from Adobe, Intel, Nvidia, Meta/Facebook, and Amazon.  MJB has financial interests in Amazon, Datagen Technologies, and Meshcapade GmbH.  While MJB is a consultant for Meshcapade, his research in this project was performed solely at, and funded solely by, the Max Planck Society.
\end{acks}

\bibliographystyle{ACM-Reference-Format}
\bibliography{main}

\end{document}